\documentclass{article} %
\usepackage{colm2024_conference}

\usepackage{booktabs}
\usepackage{graphicx}
\usepackage{enumitem}
\usepackage{wrapfig}
\usepackage{algorithm}
\usepackage{algpseudocode}
\usepackage{wrapfig}
\usepackage{float}
\usepackage{microtype}
\usepackage{amsmath}
\usepackage{amssymb}
\usepackage{colortbl}
\usepackage[utf8]{inputenc}
\definecolor{lightgray}{rgb}{0.9,0.9,0.9}
\usepackage{caption}
\usepackage{subcaption}
\usepackage{xcolor}
\usepackage{setspace}
\usepackage{url}
\usepackage{multirow}
\usepackage{colortbl}
\usepackage{tabularx}
\usepackage{blindtext}
\usepackage{pgfplots}
\pgfplotsset{compat=1.18} 
\usepackage{tikz}
\usetikzlibrary{er,positioning,bayesnet}
\usepackage{makecell}
\usepackage{tipa}
\usepackage{siunitx}
\usepackage{nicefrac}
\usepackage{tocloft}
\usepackage{listings}
\usepackage[raster,skins]{tcolorbox} %
\usepackage{xltabular}
\usepackage{adjustbox}
\usepackage{xurl}
\usepackage{multicol}

\usepackage{amsmath,amsfonts,bm}

\def\eqref#1{equation~\ref{#1}}

\def\1{\bm{1}}

\DeclareMathAlphabet{\mathsfit}{\encodingdefault}{\sfdefault}{m}{sl}
\SetMathAlphabet{\mathsfit}{bold}{\encodingdefault}{\sfdefault}{bx}{n}

\makeatletter
\DeclareRobustCommand\onedot{\futurelet\@let@token\@onedot}
\def\@onedot{\ifx\@let@token.\else.\null\fi\xspace}

\makeatother

\newcommand{\shortname}{HY3D-Bench\xspace}

\title{\shortname: Generation of 3D Assets}

\author{
\bf Tencent Hunyuan3D 
}

\begin{document}

\maketitle

\begin{figure}[h]
\centering
\includegraphics[width=0.99\linewidth]{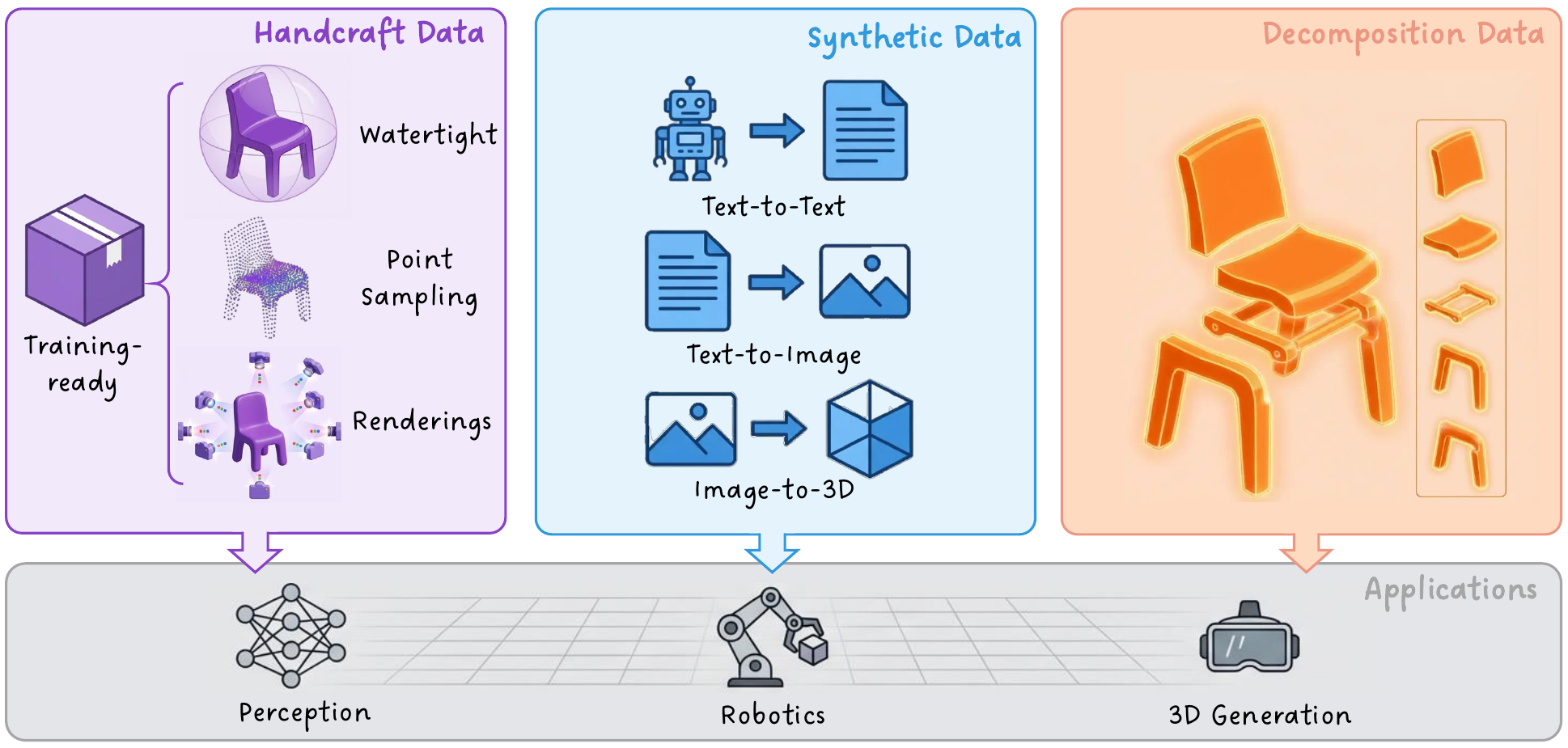}
\caption{HY3D-Bench is a unified ecosystem for high-fidelity 3D content generation. Our framework introduces (a) 252k high-quality assets with watertight meshes and multi-view renderings, (b) 240k structured part-level decomposition enabling fine-grained control, and (c) AIGC-synthesized 125k long-tail category assets. This benchmark provides standardized training data and evaluation protocols for advancing 3D generation research.}

\label{fig:teaser-top}
\end{figure}

\begin{abstract}
While recent advances in neural representations and generative models have revolutionized 3D content creation, the field remains constrained by significant data processing bottlenecks. To address this, we introduce HY3D-Bench, an open-source ecosystem designed to establish a unified, high-quality foundation for 3D generation. Our contributions are threefold: (1) We curate a library of 250k high-fidelity 3D objects distilled from large-scale repositories, employing a rigorous pipeline to deliver training-ready artifacts, including watertight meshes and multi-view renderings; (2) We introduce structured part-level decomposition, providing the granularity essential for fine-grained perception and controllable editing; and (3) We bridge real-world distribution gaps via a scalable AIGC synthesis pipeline, contributing 125k synthetic assets to enhance diversity in long-tail categories. Validated empirically through the training of Hunyuan3D-2.1-Small, HY3D-Bench democratizes access to robust data resources, aiming to catalyze innovation across 3D perception, robotics, and digital content creation.
\end{abstract}

\section{Introduction}

High-quality 3D content has become a critical asset across a broad range of fields, including 3D computer vision, generative modeling, and robotics. While pioneering large-scale repositories~\cite{objaverse,objaverseXL,wu2023omniobject3d} have provided an unprecedented volume of 3D data, their utility across these diverse fields is often hampered by significant limitations. Most raw assets in these datasets contain significant noise, non-manifold geometry, and a lack of structural granularity, which restricts their application in tasks requiring precise geometric understanding, stable generation, or complex robotic interaction.

In this work, we present \shortname, a comprehensive open-source ecosystem designed to provide a high-quality, structured, and reproducible foundation for 3D content research. Moving beyond simple mesh collection, our work integrates rigorous data engineering, standardized benchmarks, and scalable AIGC-driven synthesis to support the dual goals of 3D content understanding and creation. Our contributions are categorized into three major pillars:

First, we provide a \textbf{\textit{refined and structured 3D asset library}} with comprehensive data processing results. For each holistic object, we implement a professional pipeline to generate \textit{training-ready} assets featuring the \textit{best watertight mesh} and high-fidelity rendered images, both of which are essential for stable 3D generation training and accurate geometric perception. Crucially, we utilize a part-merging strategy to produce structured assets with consistent part-level decomposition. For the structural components, we provide the original mesh segmentation results and individual part-level watertight meshes, complemented by view-dependent RGB renderings and 2D masks for the integrated part assembly. This structural granularity provides the necessary information for part-aware generation and fine-grained perception tasks.

Second, we establish a \textbf{\textit{standardized evaluation and experiment framework}} to address the fragmentation in 3D research. We propose a rigorous benchmark comprising 400 high-quality objects across diverse categories, providing a unified platform for testing 3D generation algorithms. Unlike previous works with inconsistent evaluation protocols, we provide a complete suite of standard metrics, baselines, and a fixed experiment setting. By releasing our standardized training configurations and pre-trained model checkpoints, we empower the community to conduct fair comparisons and accelerate the rapid advancement of the 3D generation field.

Third, we introduce a \textbf{\textit{scalable AIGC-driven data synthesis pipeline}} to bridge the gap in category diversity and long-tail distribution. Recognizing that manual 3D modeling for realistic scenarios, such as shopping malls, is prohibitively expensive, we leverage the generative power of Large Language Models and Diffusion Models to synthesize diverse 3D content. Our three-step paradigm, consisting of Text-to-Text for semantic expansion, Text-to-Image for visual synthesis, and Image-to-3D for mesh reconstruction, allows us to produce a vast collection of long-tail items, allows us to produce a vast collection of long-tail items covering 20 super-categories, 130 categories, and 1,252 fine-grained sub-categories. This synthetic data provides a critical supplement for training models that can generalize to rare but crucial object categories, which is particularly vital for the robustness of generation and the diversity of robotics simulation environments.

By providing a structured, diverse, and standardized 3D content ecosystem, we aim to lower the barrier for research and drive the progress toward a unified understanding and generation of the 3D world. In summary, our contributions are as follows:
\begin{itemize}
    \item A high-quality 3D asset library featuring watertight meshes and rendered images for both holistic objects and parts.
    \item A standardized 3D benchmark and experiment framework, providing unified metrics, baselines, and model weights.
    \item An AIGC-based synthesis framework that expands 3D data diversity, focusing on long-tail assets to support broad generalization.
    \item Extensive data and infrastructure support for a wide range of downstream tasks, including 3D generation, perception pre-training, and robotics simulation.
\end{itemize}

\section{Related Work}

\begin{figure}
    \centering
    \includegraphics[width=0.9\linewidth]{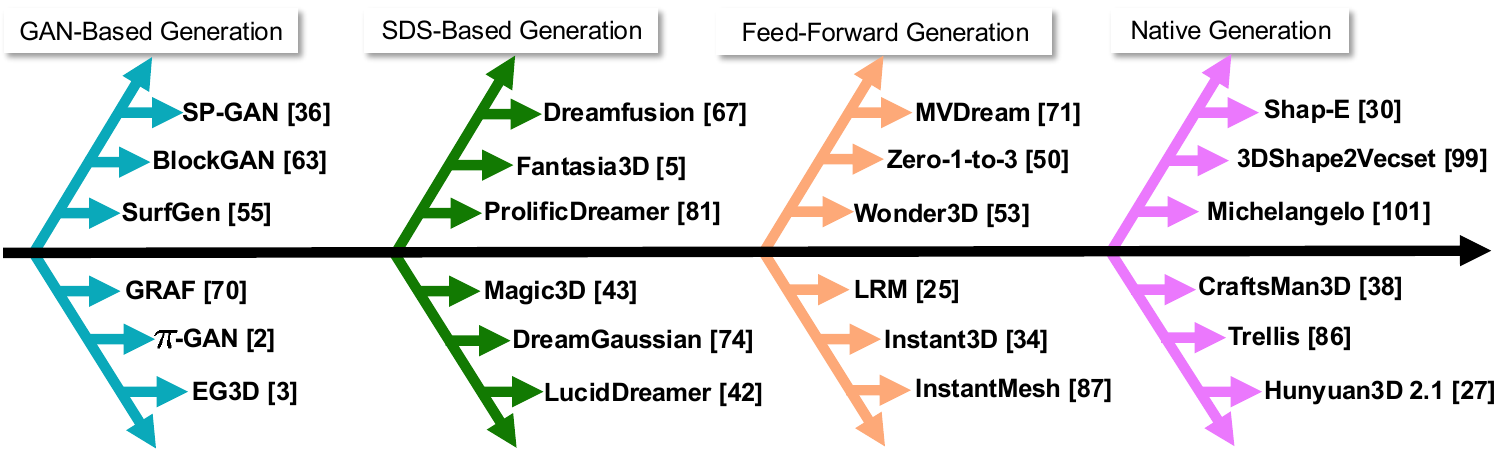}
    \caption{The evolution of the 3D generation.}
    \label{fig:related}
\end{figure}

\subsection{3D Generation}
The field of 3D generation has emerged as a cornerstone of generative AI, bridging the gap between virtual content creation and real-world applications. The evolution of 3D generation has witnessed a paradigm shift from manual modeling and scanning-based reconstruction to data-driven AI synthesis. This field can be systematically categorized into four major paradigms: GAN-based methods, SDS-based methods, feedforward-based methods, and 3D native generation, as shown in Figure~\ref{fig:related}.

{\bf GAN-based generation.} Generative Adversarial Networks (GANs)~\cite{goodfellow2020generative} established the initial paradigm for high-fidelity synthesis by optimizing a minimax objective between a generator and a discriminator. 
Following this, early works~\cite{luo2021surfgen,chen2022gdna} based on explicit representations, such as voxel grids~\cite{wu2016learning,nguyen2020blockgan,nguyen2019hologan} and point clouds~\cite{shu20193d,li2021sp}, attempt to generate 3D shapes directly, yet methods often suffer from cubic memory complexity and limited resolution. 
The advent of Neural Radiance Fields (NeRF)~\cite{mildenhall2021nerf} shifted the focus toward 3D-aware image synthesis, where models are trained on multi-view 2D images to learn underlying 3D geometry. 
Seminal works such as GRAF~\cite{schwarz2020graf} and $\pi$-GAN~\cite{chan2021pi} integrated conditional radiance fields with adversarial training, utilizing coordinate-based MLPs to enforce multi-view consistency. 
However, fully implicit backbones proved computationally expensive for high-resolution rendering. 
Addressing this, EG3D~\cite{chan2022efficient} proposed a hybrid explicit-implicit tri-plane representation, leveraging the efficiency of StyleGAN2~\cite{karras2020analyzing} to generate feature planes that are subsequently decoded by a lightweight MLP via volume rendering. 
While GANs achieve rapid inference speeds, they remain prone to training instability and mode collapse, particularly when scaling to diverse, open-domain datasets.

{\bf SDS-based generation.} 
The scarcity of large-scale, annotated 3D datasets has historically hindered the development of generative 3D models compared to their 2D counterparts. To circumvent this data bottleneck, recent approaches have shifted towards optimization-based pipelines that leverage pre-trained 2D text-to-image diffusion models as strong priors. The pioneer work, DreamFusion~\cite{poole2022dreamfusion}, introduced Score Distillation Sampling (SDS), a method that optimizes a differentiable 3D representation—typically a NeRF—such that its rendered views maintain high likelihood under a frozen 2D diffusion model. By replacing the standard diffusion denoising loss with a gradient-based score matching objective, SDS enables the distillation of semantic knowledge from 2D foundation models into consistent 3D structures without requiring 3D ground truth. Following DreamFusion, subsequent works~\cite{wang2023prolificdreamer,EnVision2023luciddreamer,fantasia3d,lin2023magic3d,sweetdreamer} are proposed to further enhance the quality of 3D generation. More recently, the field has transitioned from implicit NeRF representations to explicit 3D Gaussian Splatting \cite{kerbl20233d} to achieve real-time rendering and improved convergence speeds. Methods such as DreamGaussian~\cite{tang2023dreamgaussian} and GaussianDreamer\cite{yi2024gaussiandreamer} adapt SDS to optimize 3D Gaussian parameters, significantly reducing generation time while maintaining visual fidelity. However, these SDS-based methods usually suffer from the "Janus problem" (multi-face artifacts) due to the lack of explicit 3D geometry-related constraints. In addition, the problem of long-term optimization also poses real-time challenges for such methods.

{\bf Feedforward generation.} In contrast to optimization-based paradigms that require computationally expensive per-instance training (e.g., via Score Distillation Sampling), feedforward methods prioritize inference efficiency by learning a direct, amortized mapping from input prompts to 3D representations. A foundational direction in feedforward generation follows two stages: multi-view (MV) image synthesis and 3D reconstruction. For instance, MVDream~\cite{shimvdream} introduces a multi-view diffusion model conditioned on camera poses, enabling the generation of geometrically consistent MV images from text, which are then fed into a neural radiance field (NeRF) or mesh reconstruction pipeline to yield 3D assets. Subsequent methods~\cite{liu2023zero,long2024wonder3d,li2023instant3d,liu2023one,liu2023syncdreamer,li2024era3d} attempt to improve the multi-view consistency and image resolution to obtain high-quality 3D assets. Another prominent yet distinct feedforward paradigm for 3D generation is the Large Reconstruction Model (LRM) approach~\cite{hong2023lrm}. LRM aims to learn a universal reconstruction capability from large-scale 3D data, enabling it to generate 3D representations directly from textual or sparse visual inputs through a single forward pass. These models leverage the richness of large-scale 3D datasets to learn generalized 3D shape priors, which are then used to amortize the optimization cost across multiple generation tasks. Building on the LRM paradigm, subsequent works~\cite{wang2023pf,tang2024lgm,xu2024instantmesh,li2023instant3d} have proposed targeted improvements to enhance generation quality, efficiency, and generalization. For instance, LGM~\cite{tang2024lgm} proposes a representation based on Gaussian features to improve the resolution of 3D models. While these feedforward methods outperform optimization-based counterparts in inference speed and geometric quality, they are constrained by the resolution of 2D images and the lack of learning and understanding of the spatial distribution of 3D data, making it challenging to generate fine-grained and accurate 3D geometries.

{\bf Native generation.} 
Unlike 2D-lifting approaches, native 3D generation methods directly learn the 3D representations, such as point clouds~\cite{zhou20213d,pointflow,nichol2022point,luo2021diffusion}, meshes~\cite{nash2020polygen,jun2023shap,Liu2023MeshDiffusion}, and implicit functions~\cite{chen2019learning,park2019deepsdf}, from large-scale 3D assets, typically yielding superior geometric consistency and topology. However, these non-compressed methods are typically constrained by computational complexity and resolution, making it challenging to generate high-quality geometries. A pivotal breakthrough in native 3D generation came with large-scale 3D datasets~\cite{objaverse,objaverseXL} and 3DShape2Vecset~\cite{zhang20233dshape2vecset}, which innovatively adopted the processing paradigm of 2D Stable Diffusion and constructed a 3D VAE (Variational Autoencoder) to compress 3D shapes into compact VecSet representation. With this representation, 3DShape2Vecset enabled the construction of diffusion models for both conditional and unconditional 3D generation. Following this workflow, subsequent works~\cite{zhao2024michelangelo,zhang2024clay,li2024craftsman,li2025triposg,wu2024direct3d,chen2025dora,hunyuan3d2025hunyuan3d} strive to enhance the model's generalization ability and geometric fidelity by scaling up the model and data. In contrast to implicit VexSet, several approaches~\cite{ren2024xcube,xiang2024structured,he2025sparseflex,wu2025direct3d,huang2025spar3d} apply structured voxel-based representation to preserve spatial structure in latent space. Additionally, there are a series of studies~\cite{dong2025crossgen,ye2025nano3d} that apply native generation on other specific issues. For example, PoseMaster~\cite{yan2025posemaster} and Hunyuan3D-Omni~\cite{hunyuan3d2025omni} introduce a native controllable generation model to achieve control on point, voxel, bounding box, and skeleton. There is also a line of research on fine-grained 3D generation, whose primary goal is to produce part-aware results. One set of works\cite{yang2025holopart,liu2023partslip,ma2025p3,kim2024partstad,zhong2024meshsegmenter,abdelreheem2023satr,tang2024segment,thai20243x2,xue2023zerops,yang2024sampart3d,liu2024part123, deng2025geosam2,zhou2024point,fischer2024sama,ma2025find,liu2025partfield,zhu2025partsam,paul2025name,li2025auto} adopts a segmentation-based pipeline: starting from a holistic object and decomposing it into parts via segmentation. Another set of works\cite{yan2025x,li2025moca,yang2025omnipart,lin2025partcrafter,tang2025efficient,dong2025one,ding2025fullpart,he2025unipart,yang2025partdiffuser} instead follows a part-aware generation paradigm, directly generating 3D objects with explicit part structures. In addition, some works~\cite{chang2015shapenet,mo2019partnet,collins2022abo,wang2025partnext,dong2025one,yang2024sampart3d,geng2023gapartnet,deng20213d} provide datasets with part-level annotations. While these methods present impressive performance in the 3D generation task, they usually rely on high-quality data processing in terms of the part-aware mesh and watertight mesh. In this paper, we open-source large-scale processed data that can be used to train 3D VAE and diffusion directly.


\subsection{3D Datasets}
The advancement of 3D generation models is inherently tied to the availability of high-quality benchmark datasets, which provide the foundational data support for model training, validation, and evaluation. Early 3D benchmark datasets~\cite{zhou2016thingi10k,fu20213d,downs2022google,wu2023omniobject3d}, such as ShapeNet~\cite{chang2015shapenet}, laid the initial groundwork for the development of 3D generation research. However, these datasets suffer from limitations such as a limited number of categories, simple geometry, and small quantities, which severely constrain the generalization capabilities of trained 3D generation models. This bottleneck has long restricted the further advancement of 3D generation technology towards more practical and versatile scenarios. The emergence of large-scale 3D datasets with complex geometric structures has broken this deadlock, among which Objaverse stands out as a pivotal milestone. As the large-scale, diverse 3D object dataset, Objaverse~\cite{objaverse} and Objaver-XL~\cite{objaverseXL} contain millions of 3D models spanning a wide range of categories, including complex geometric structures such as articulated objects, organic shapes, and detailed industrial parts. The release of Objaverse has significantly empowered the development of 3D generation technology, particularly fostering the advent of a new generation of large-scale 3D generation models. 


However, a critical challenge persists in the current 3D generation research landscape: mainstream 3D generation models typically require extensive preprocessing of raw 3D data to generate task-specific representations, such as rendered images, watertight meshes, and corresponding Signed Distance Function (SDF) fields. This preprocessing step not only increases the entry barrier for researchers new to 3D generation, requiring proficiency in specialized data processing techniques, but also imposes substantial computational burdens. Although open-source data processing scripts have been developed to alleviate some of these difficulties by automating certain preprocessing workflows, processing large-scale training datasets (often involving millions of 3D models) demands enormous GPU and CPU computational resources. This resource-intensive preprocessing requirement remains a significant bottleneck for the broader research community, hindering the rapid iteration and widespread adoption of 3D generation models.

To address this critical challenge, in this paper, we directly provide a high-quality dataset of 200k samples specifically tailored for training 3D Variational Autoencoders (3D VAE) and 3D diffusion models. The data samples are curated from two large-scale 3D repositories, Objaverse and Objaverse-XL, ensuring rich category diversity and complex geometric characteristics. Notably, we process the 3D meshes to obtain watertight meshes at a resolution of 512, which effectively preserves a large number of fine-grained details from the original meshes. By offering this preprocessed, high-resolution 3D dataset, we aim to reduce the computational and technical burdens on researchers, lower the entry barrier for 3D generation research, and further facilitate the advancement of the field.

\section{Methods}


{\bf VAE}. Given an input point cloud $P \in R^{N \times (3 + C)}$ sampled from the mesh surface, where $C$ denotes surface normals, 3D VAE first extract point features and then obtain the corresponding latent vector set $Z \in R^{L \times d}$ via resampling from estimated distribution, where $L$ and $d$ indicate the length and dimension of latent VecSet, respectively. Subsequently, a decoder is applied to reconstruct the signed distance function (SDF) field $F_{sdf}$, in which we can leverage the iso-surface extraction to obtain explicit mesh output. The procedure of VAE can be formulated as follows:
\begin{align}
Z = \mathcal{E}(P), F_{sdf} = \mathcal{D}(Z)
\end{align}


{\bf Diffusion}. Given an image and its latent set representation $Z$ of a shape, the 3D diffusion model aims to model the denoising process, thereby achieving conditional generation from an arbitrary image. It first leverages an image encoder, such as DINO-v2~\cite{}, to capture image embeddings $c_i$ and then exploits the multi layers of DiT to predict the added noise or velocity. For a flow matching model used in Hunyuan3D 2.1~\cite{}, its training objective is to transform a simple noise distribution $x_0 \sim \mathcal{N}(0, I)$ into a complex data distribution $x_1 \sim D$ conditioned on image embeddings $c_i$, which can be formulated as follows:
\begin{equation}
\mathbb{E}_{t, {x}_0, {x_1}, c}\vert\vert{v}_\theta({x}, t, c)-(x_1-x_0)\vert\vert_2^2
\end{equation}




\section{Hunyuan Objarverse}

The currently open-source Objaverse series datasets~\cite{objaverse,objaverseXL} contain a vast collection of raw 3D assets available for access and download. However, these raw assets suffer from numerous critical issues that urgently need to be addressed, rendering them unsuitable for direct application in downstream tasks such as 3D generation.

First, from a technical specification perspective, various types of 3D assets produced by different 3D modeling software (such as Blender, Maya, 3ds Max, etc.) exhibit significant format discrepancies and lack of standardization. Specifically: (1) Inconsistent coordinate system definitions: Different software packages adopt varying coordinate system conventions (e.g., left-handed vs. right-handed systems, Y-up vs. Z-up, etc.), resulting in orientation errors or mirror flipping when assets are loaded in different environments; (2) Complex and diverse asset construction methods: Many assets employ multi-level node hierarchies, contain parent-child node scale inheritance relationships, and include hidden transformation matrices, which greatly increase the complexity of data processing.

Second, from a data quality perspective, the quality of various 3D assets is highly inconsistent, exhibiting significant heterogeneity. The main issues include: (1) Poor geometric quality: A large number of assets have overly simplified meshes with insufficient polygon counts, failing to accurately represent the detailed features of objects. Additionally, severe topological defects exist, such as non-manifold edges, self-intersecting faces, and isolated vertices. These problems render the assets unsuitable for tasks requiring watertight meshes (such as physical simulation, 3D printing, etc.); (2) Texture mapping errors: Some assets have serious UV unwrapping problems, with incorrect texture-to-geometry mapping, excessively low texture resolution, or missing textures, which compromise rendering quality.

Finally, from a data ecosystem perspective, existing datasets also suffer from the following systemic deficiencies: (1) Severely imbalanced category distribution: The datasets exhibit pronounced long-tail distribution characteristics, with abundant assets in common categories (such as chairs and tables), while assets in many rare categories that are important for real-world applications are extremely scarce, limiting the generalization capability of models; (2) Lack of structured information: The vast majority of assets are holistic, monolithic meshes, lacking hierarchical part decomposition and assembly relationship descriptions, which severely constrains the development of advanced applications such as fine-grained understanding, editable generation, and robotic manipulation.

To address the above issues, we first process and clean the raw 3D assets from the Objaverse series datasets through a combination of automated processing and manual assistance, obtaining a collection of high-quality static mesh processing results. We hope that researchers can conduct algorithmic exploration and research on a unified, high-quality benchmark. Second, we further perform part-level processing to obtain structured assets with consistent part-level decomposition, yielding a batch of high-quality original mesh segmentation results and individual part-level watertight meshes. We hope that researchers can further pursue more fine-grained algorithmic exploration and research. Finally, based on real-world object and product categories, we generate a collection of category-balanced 3D assets, aiming to help improve the generalization capability and algorithmic exploration of downstream tasks such as grasping.

\subsection{Existing Enhanced Objaverse Dataset}




Objaverse~\cite{objaverse} and Objaverse-XL~\cite{objaverseXL} provide the 3D research community with ultra-large-scale, diverse 3D asset datasets. However, researchers have been consistently challenged by issues of inconsistent data quality and complex 3D data processing workflows. Multiple subsequent works~\cite{lin2025objaverseplusplus, lu2025objaverseoa, jin2025canoobjdataset, qian2024objaversemix} have approached the filtering, processing, and enhancement of Objaverse from different perspectives.

From the perspective of data quality filtering, Objaverse++~\cite{lin2025objaverseplusplus} manually annotated 10,000 samples as training data based on key quality dimensions such as model transparency, single-object completeness, and scene attributes. Subsequently, a specialized quality assessment model was trained to ultimately filter out 500,000 high-quality samples.

From the perspective of geometric normalization, Objaverse-OA~\cite{lu2025objaverseoa} and Canonical Objaverse Dataset~\cite{jin2025canoobjdataset} addressed the critical issue of inconsistent 3D model orientations. Objaverse-OA established orientation normalization standards by annotating 14,000 orientation-aligned samples, while Canonical Objaverse Dataset utilized automated methods to annotate 32,000 samples.

From the perspective of data representation diversity, Objaverse-MIX~\cite{qian2024objaversemix} provided large-scale processing results containing 900,000 samples, offering multiple geometric representations such as point clouds, meshes, and voxels for each asset, accompanied by rendered images and text annotations, constructing a relatively complete training data asset package.

In summary, although existing works have improved Objaverse from different dimensions including quality filtering, orientation normalization, and multi-modal representation, the following systemic deficiencies still exist: (1) Insufficient comprehensiveness and depth in data processing, lacking a complete processing pipeline that covers format standardization, topology repair, high-quality rendering, and diverse sampling; (2) Filtered data still requires complex workflows before it can be used for training; (3) Processed data struggles to meet the training requirements of current 3D generation models due to issues such as fixed rendering viewpoints and single point cloud sampling strategy. In contrast, we employ a complete processing pipeline to curate high-quality data and process them into training-ready asset packages for researchers to use. Furthermore, we generate a collection of high-quality product assets as a supplement.


\subsection{Full-level Data Processing}
\begin{figure}[h]
    \centering
    \includegraphics[width=0.98\linewidth]{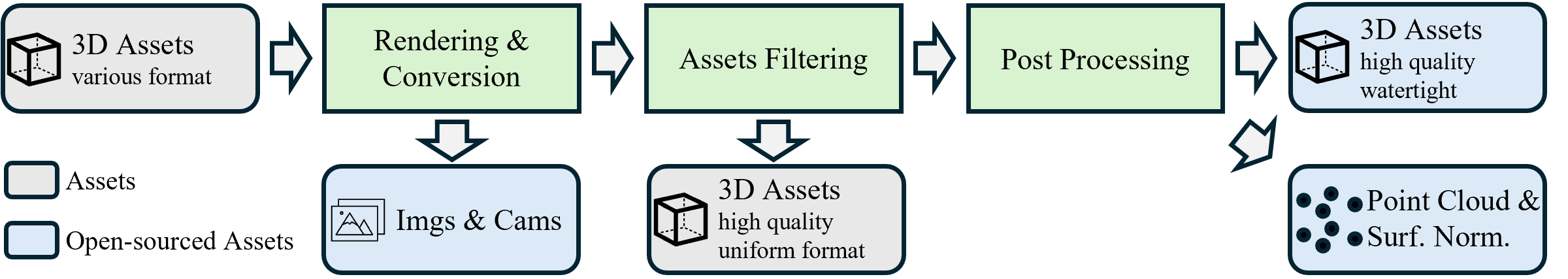}
    \caption{Full-level Data Processing Pipeline.}
    \label{fig:full_level_data_pipeline}
\end{figure}

For full-level data, our processing pipeline consists primarily of three core steps: data rendering and format conversion, asset filtering, and post processing. Through this systematic processing workflow, we are able to obtain a collection of high-quality, training-ready data for static 3D generation networks. The overall processing pipeline is illustrated in Figure~\ref{fig:full_level_data_pipeline}, with each step carefully designed to ensure the quality and consistency of the output data.

\textbf{Data Rendering and Conversion.} Considering the diverse sources and varied formats of 3D assets in the original Objaverse dataset~\cite{objaverse,objaverseXL}, we first need to establish a unified data standard. 
We combine manual annotation and automated conversion workflows to uniformly convert all 3D assets into single-frame static mesh representations with aligned orientations.
This standardization step is crucial for subsequent processing, as it eliminates coordinate system differences between different modeling software and excludes multi-view rendering inconsistencies caused by model animations. Subsequently, we use Blender as the rendering engine to perform multi-view rendering of each standardized static mesh. The rendering configuration includes two camera modes—orthographic projection and perspective projection—to cover different visual representation requirements. Finally, we uniformly export and store the processed static meshes in PLY format, which offers excellent cross-platform compatibility and efficient storage characteristics.

\textbf{Assets Filtering.} To ensure the high quality of training data, we establish a rigorous multi-dimensional filtering criteria. We comprehensively utilize the visual quality of rendering results and the geometric attributes of original 3D assets for data  filtering, primarily excluding the following three categories of inadequate data:
(1) Data with poor geometric quality. The original assets contain a large number of duplicated and overly simplified 3D assets. These assets typically exhibit: extremely low polygon counts, lack of necessary geometric details, and overly simple topological structures. We identify and exclude such assets by setting polygon count thresholds and calculating geometric complexity metrics. Retaining geometrically rich meshes with sufficient details can provide more valuable learning signals for the model.
(2) Data with poor texture quality. Texture quality directly affects the visual performance of rendering results. We exclude data with low image rendering quality due to the following reasons: serious UV mapping problems; overlapping faces in the geometry causing abnormal texture display; excessively low texture resolution or missing texture maps, etc.
(3) Data with large areas of thin structures. Thin structures pose special challenges in 3D generation tasks. We choose to exclude such data based on two main considerations: On one hand, from the perspective of implicit representations, the Signed Distance Field (SDF) at thin structures undergoes abrupt jumps, transitioning from positive to negative values within an extremely small spatial range, which significantly increases the difficulty of model learning and fitting and can easily lead to training instability; On the other hand, from the perspective of multi-view consistency, thin structures under certain viewpoints are difficult to observe or even completely invisible in 2D images (such as when viewing along the thin sheet direction), which reduces the stability and convergence speed of model learning. Therefore, excluding assets containing large areas of thin structures helps improve overall training effectiveness.

\textbf{Post Processing}. 
After obtaining high-quality 3D assets, we further perform post-processing to generate training-ready data. The post-processing steps primarily include: watertight processing and point cloud sampling. (1) Watertight processing. Given an artist-created triangle mesh, we first compute the Unsigned Distance Field (UDF) on a uniform grid with $512^3$ resolution, and extract an $\epsilon$-contour thin shell mesh $\mathcal{M}$ using Marching Cubes with $\epsilon=1/512$. We then sample points on $\mathcal{M}$ and apply Delaunay triangulation to construct a volumetric tetrahedral mesh. Following the approach of ConvexMeshing~\cite{diazzi2021convexmeshing}, we optimize the tetrahedral cell labels (0 for inner, 1 for outer) using graph cut optimization, and extract the boundary surface as the final watertight mesh. (2) Point cloud sampling. Following the sampling strategies of Dora~\cite{chen2025dora} and Hunyuan3D 2.1~\cite{hunyuan3d2025hunyuan3d}, we implement a hybrid sampling scheme on watertight meshes by combining surface uniform sampling and edge importance sampling to ensure that the sampled point clouds can both adequately represent the overall geometric shape and accurately capture local detail features. It is worth noting that we rotated the coordinate system to Y-up during the post-processing stage.



\subsection{Part-level Data Processing}

\begin{figure}[h]
    \centering
    \includegraphics[width=0.98\linewidth]{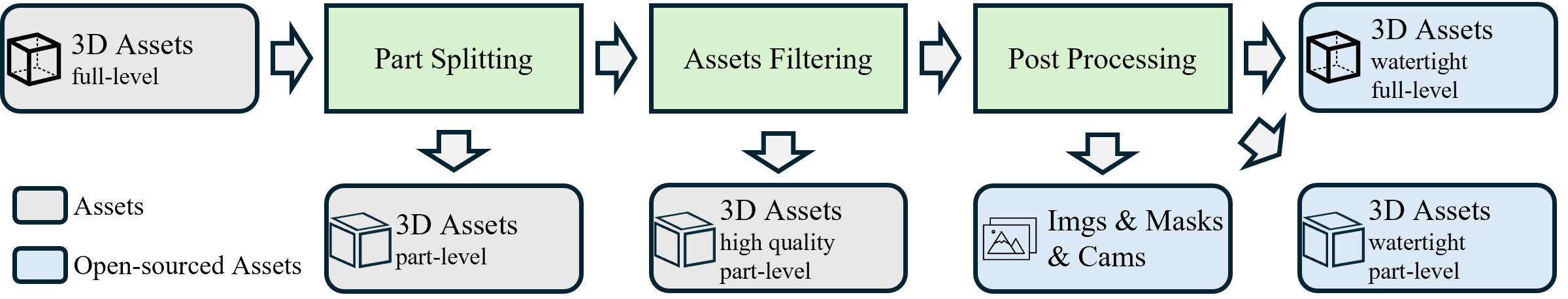}
    \caption{Part-level Data Processing Pipeline.}
    \label{fig:part_level_data_pipeline}
\end{figure}

For part-level data, we have designed a specialized data processing pipeline aimed at decomposing holistic static meshes into semantically consistent component collections. This pipeline consists primarily of three core steps: part splitting, asset filtering, and post-processing. Through this systematic processing workflow, we are able to transform original holistic static meshes into part-level components suitable for training part-aware generation networks. The overall processing pipeline is illustrated in Figure~\ref{fig:part_level_data_pipeline}.

\textbf{Part Splitting.} Part splitting is the critical step of breaking down holistic meshes into meaningful part units. We adopt a splitting strategy based on topological connectivity, first utilizing Connected Component Analysis to perform initial splitting of 3D assets, obtaining a collection of topologically independent original components. This step can automatically identify physically separated parts within the mesh, aligning the division of the holistic mesh with the semantic granularity designed by artists during the creation process.

After obtaining the initial decomposition results, we need to perform preliminary quality control filtering to exclude two types of extreme cases: (1) Complex assets with excessive components (component count $\textgreater$ 888): These assets typically contain numerous trivial small parts or decorative elements, whose excessive complexity significantly increases the difficulty of data processing; (2) Indivisible assets (component count $\textless$ 2): These assets cannot provide structural information at the part-level and do not meet the data requirements of part-aware generation tasks and are therefore excluded.

To address the over-fragmentation issue in the initial decomposition results, we further implement an automatic merging strategy. Specifically, we calculate the surface area of each original component and set area thresholds to identify small trivial parts. For components with areas significantly below the threshold, we merge them into adjacent larger components based on spatial adjacency relationships, thereby obtaining more reasonable part granularity. After this merging process, the final component count for the vast majority of assets is controlled between 10 and 40, a range that both retains sufficient semantic granularity and avoids excessive complexity, making it highly suitable for the training requirements of part-level generation tasks.

\textbf{Asset Filtering.} After completing part splitting, we establish a rigorous set of multi-dimensional filtering criteria to ensure that the retained data possesses both reasonable part-level structure and is suitable for model learning. The specific filtering rules are as follows: (1) Component quantity reasonableness verification. We exclude data with too few components ($\leq$1) or too many components ($\textgreater$50). Too few components indicate splitting failure or that the asset itself lacks structural complexity; too many components suggest that even after merging, the asset remains overly complex and may affect network learning. This filtering ensures that all assets in the dataset have moderate part-level complexity. (2) Part scale balance verification. We exclude data where the area of a single component exceeds 85\% of the total area of the surface of the object. These assets have extremely imbalanced part distributions, typically manifesting as one massive dominant part accompanied by several tiny auxiliary parts (such as a large tabletop with tiny leg connectors). This imbalanced part distribution is detrimental to the model learning reasonable proportional relationships and compositional logic among parts, and is therefore excluded. (3) Isolated small part quantity verification. We exclude data containing too many isolated small-area components. These isolated small parts are often decorative trivial elements that typically do not provide valuable semantic information and can interfere with the model's learning of relationships among major parts. By counting the proportion of isolated small parts, we can effectively identify and filter such low-quality data.

\textbf{Post Processing}. The post processing step aims to generate complete training data packages for each asset that passes the filtering. First, we perform systematic multi-view rendering of assets based on splitting results, generating two types of complementary image data: (1) RGB texture images: Rendered using original texture maps to obtain realistic appearance representations; (2) Part ID masks: Based on the splitting results, we assign a unique ID to each part and render 2D part mask images. In the mask images, each pixel's value corresponds to the ID of the part it belongs to. By simultaneously providing RGB images and part masks, this data can be used for training controllable part-aware object generation model. Subsequently, we perform watertight processing on the geometric data, separately processing the holistic mesh and individual part meshes: (1) Holistic mesh watertightening: We perform watertight processing on the merged complete object mesh to generate a topologically closed holistic representation; (2) Part mesh watertightening: We perform watertight processing on each independent part mesh separately, ensuring that each part is a topologically closed geometric entity. This step is crucial because many parts may have open boundaries at connection points after decomposition, and watertight processing can complete these boundaries, making each part an independent, complete 3D object.

Through the above complete processing pipeline, we ultimately obtain a high-quality part-level 3D dataset, with each sample containing: a reasonable number of semantically consistent parts, multi-view RGB images and part masks, and watertight holistic and part meshes. This rich data lays a solid foundation for training powerful part generation models, fine-grained 3D understanding or editing models, and simulation environments supporting complex robotic manipulation.

\subsection{Synthetic Data Generation}

\begin{figure}[h]
    \centering
    \includegraphics[width=0.8\linewidth]{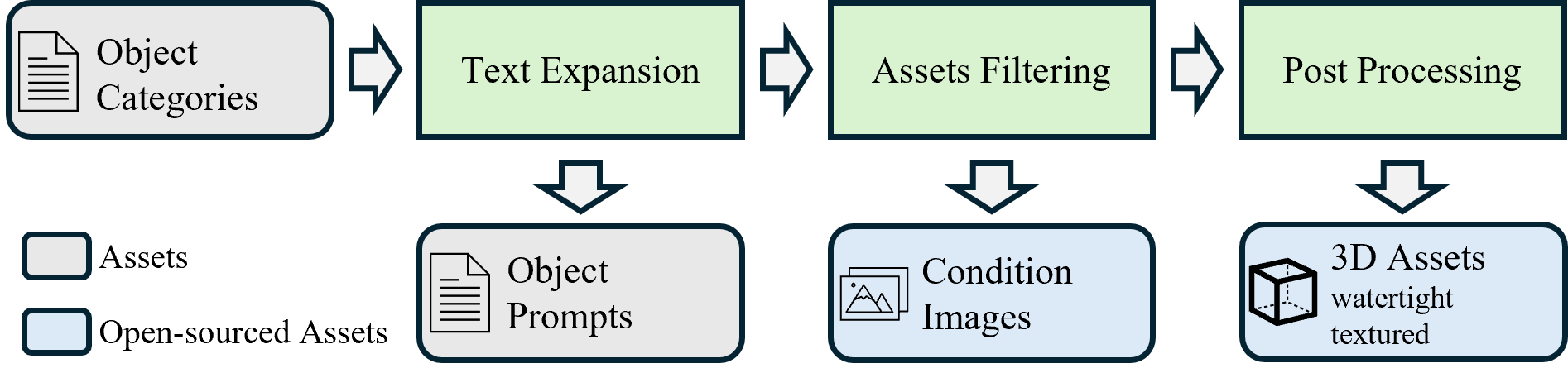}
    \caption{Synthetic Data Generating Pipeline.}
    \label{fig:synthetic_data_pipeline}
\end{figure}

We leverage the powerful priors of generative models to synthesize data, aiming to bridge the significant gap in sample counts across object categories that exists in real-world datasets. To achieve this goal, our data synthesis pipeline consists of three main steps: text expansion, image generation, and 3D generation. The overall pipeline is illustrated in the Figure~\ref{fig:synthetic_data_pipeline}.

\textbf{Text Expansion}. We first collected and organized a complete e-commerce product category system from mainstream e-commerce platforms and product databases, constructing a category hierarchy that comprehensively covers real-world products. After excluding service-oriented virtual products (such as insurance, membership services.), we ultimately retained 1,252 specific physical product categories.

Using these product categories as semantic conditions, we employ an LLM model to generate detailed and diverse product descriptions. Our prompt design is centered around the following three points: (1) Ensuring basic rationality and authenticity, generating physically and logically reasonable descriptions around the category; (2) Providing rich visual details, including key attributes such as the object's shape, material, color, and size proportions; (3) Expanding diversity, imaginatively expanding the product's form, materials, and other content within a reasonable range, setting aside limitations of actual craftsmanship, cost, and other factors.


\textbf{Image Generation}. We select Qwen-Image to transform text descriptions into images. Although this model performs excellently in text understanding and image quality, as a general-purpose text-to-image model, it often generates images containing complex backgrounds, or viewpoints unsuitable for 3D generation. To ensure that the generated images are suitable for subsequent 3D generation step, we customize the model behavior through LoRA fine-tuning.

Specifically, our fine-tuning objective is to enable the model to generate images that meet the following quality standards: (1) Clean background: Solid color or simple gradient backgrounds with no complex scene elements, facilitating the separation of foreground objects; (2) Complete object: Ensuring that the overall geometric features can be accurately captured; (3) Appropriate position: The object is located at the image center, occupying a suitable proportion of the frame, avoiding being too large or too small; (4) Reasonable view point: Adopting three-quarter views or other information-rich observation angles that can simultaneously display multiple faces of the object, providing sufficient geometric cues for 3D generation; (5) Information-rich: Clearly displaying the object's key structural features, material properties, and detail elements.


\textbf{3D Generation}. We select the industry-leading HY3D-3.0 model~\cite{hunyuan3d_online} as our 3D generation engine. Leveraging the powerful capabilities of the HY3D-3.0 model, we are able to obtain high-quality 3D assets with the following characteristics: (1) Fine geometry: The generated meshes possess rich geometric details, accurately reconstructing the object's shape features, including complex structures such as edges, bumps, and holes; (2) Clear textures: Accurate texture mapping, with visual attributes such as color, material, and surface details highly consistent with the input image.


\subsection{Data Distribution and Visualization}




{\bf Full-level Data}. Using Objaverse and Objaverse-XL as base data sources, we conducted rigorous quality filtering and data processing workflows, ultimately curating 252,676 high-quality 3D assets for in-depth processing. These assets have undergone the complete data processing pipeline described above, ensuring that each asset meets training-ready standards.

To support model training and scientific evaluation, we perform a split of the dataset: 252,000 samples are allocated to the training set for comprehensive model learning; 276 samples are allocated to the validation set for hyperparameter tuning and model selection during training; and 400 samples are allocated to the test set for final model performance evaluation and benchmarking.

In terms of category coverage, the entire dataset spans 19 top-level categories, further subdivided into 74 mid-level subcategories, and ultimately contains 389 fine-grained classifications, such as Animal-Virtual/Extinct Animals-Anthropomorphic Animals, Weapon-Firearms-Guns.

\begin{figure}[h]
    \centering
    \includegraphics[width=0.9\linewidth]{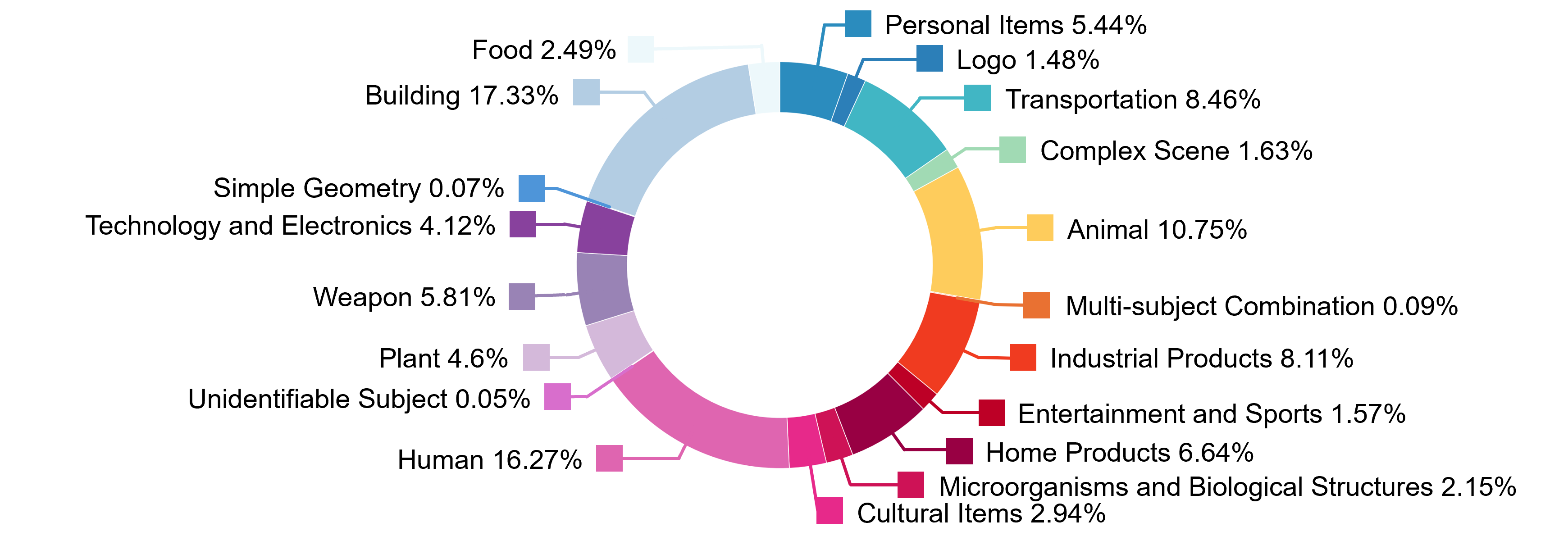}
    \caption{The Top-level Category Distribution of Full-level Data.}
    \label{fig:top_level_distribution_full_data}
\end{figure}

\begin{figure}[h]
    \centering
    \includegraphics[width=0.82\linewidth]{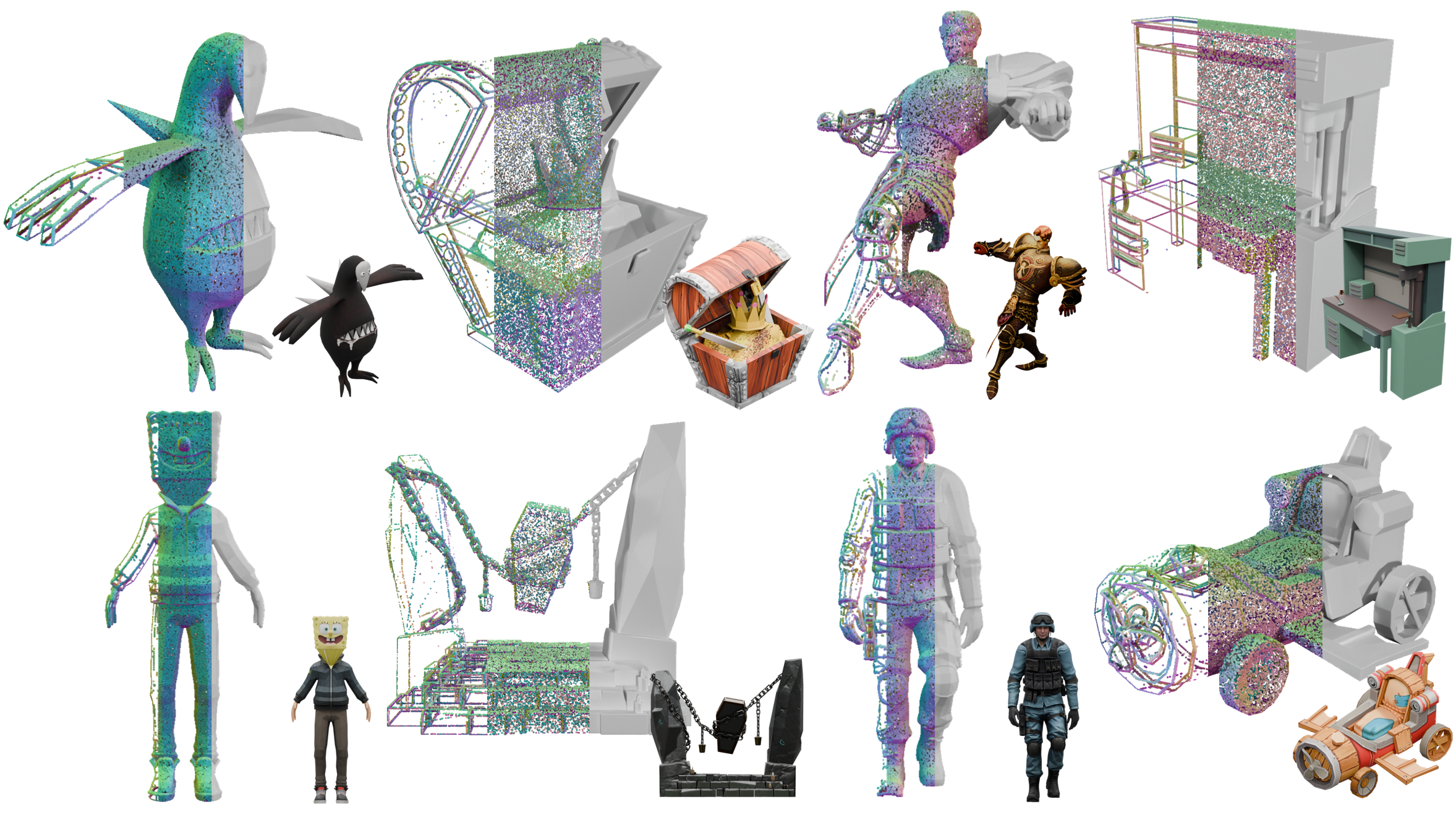}
    \caption{Visualization of the full-level dataset, including sharp edge point clouds, random surface point clouds, watertight meshes, and rendered images.}
    \label{fig:full_data_vis}
\end{figure}



\begin{figure}[h]
    \centering
    \includegraphics[width=0.95\linewidth]{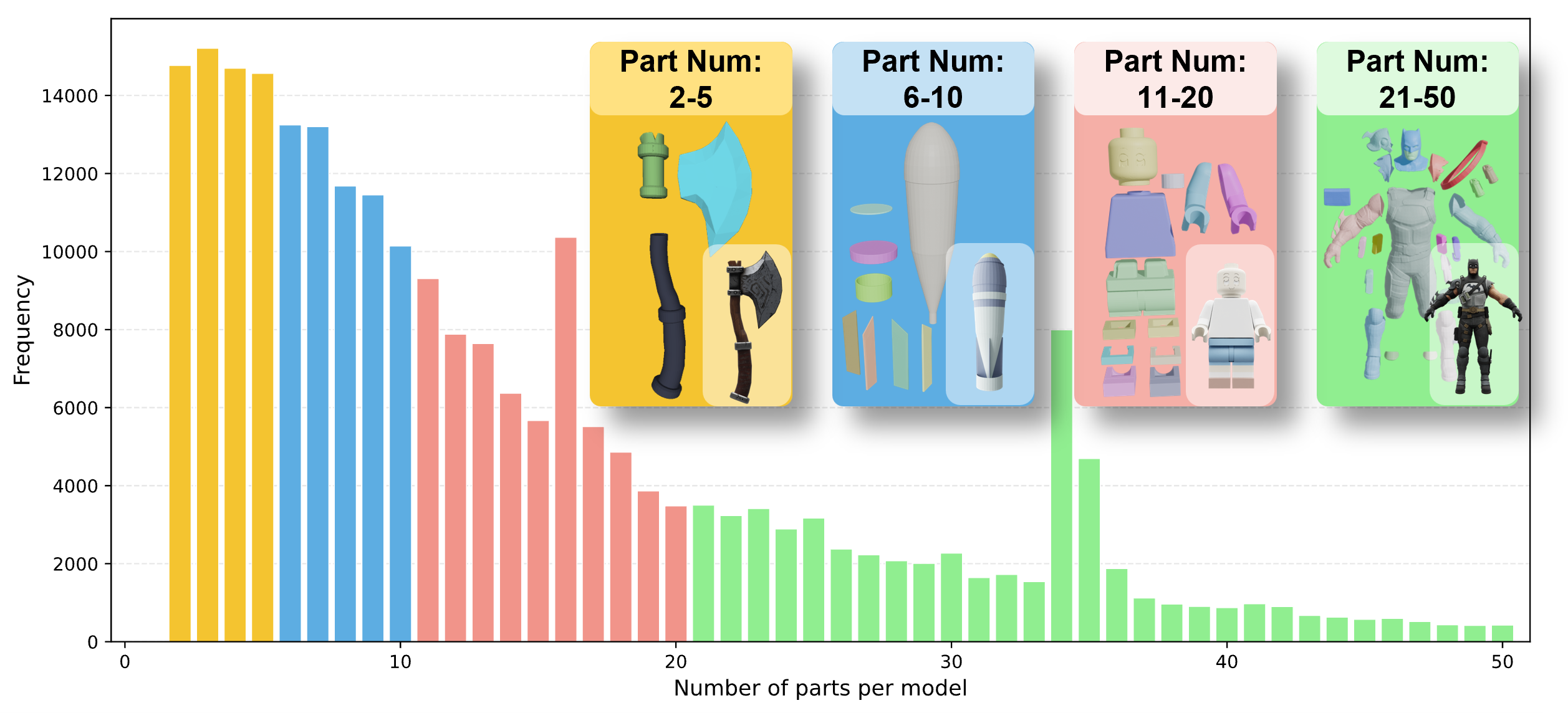}
    \caption{The Component Distribution of Part-level Data.The prominent peaks at 16, 34, and 35 primarily stem from humanoid models that share identical geometric structures but differ in texture. Considering that various research scenarios and application needs may require such texture variant data, we chose to retain this portion of the data without deduplication.}
    \label{fig:component_distribution_part_data}
\end{figure}

{\bf Part-level Data}. The part-level dataset comprises 240,524 samples in total, with a mean component count of 14.13 and a median of 11, exhibiting a diverse distribution of component complexity. Specifically, 24.63\% of samples contain 2-5 components, representing relatively simple object structures; 24.83\% of samples contain 6-10 components, covering objects with moderate structural complexity; 27.00\% of samples contain 11-20 components, encompassing more intricate assemblies; and the remaining samples contain 21-50 components, representing highly complex multi-part objects. The detailed statistical distribution of component counts is illustrated in Fig.~\ref{fig:component_distribution_part_data}.

\begin{figure}[h]
    \centering
    \includegraphics[width=0.95\linewidth]{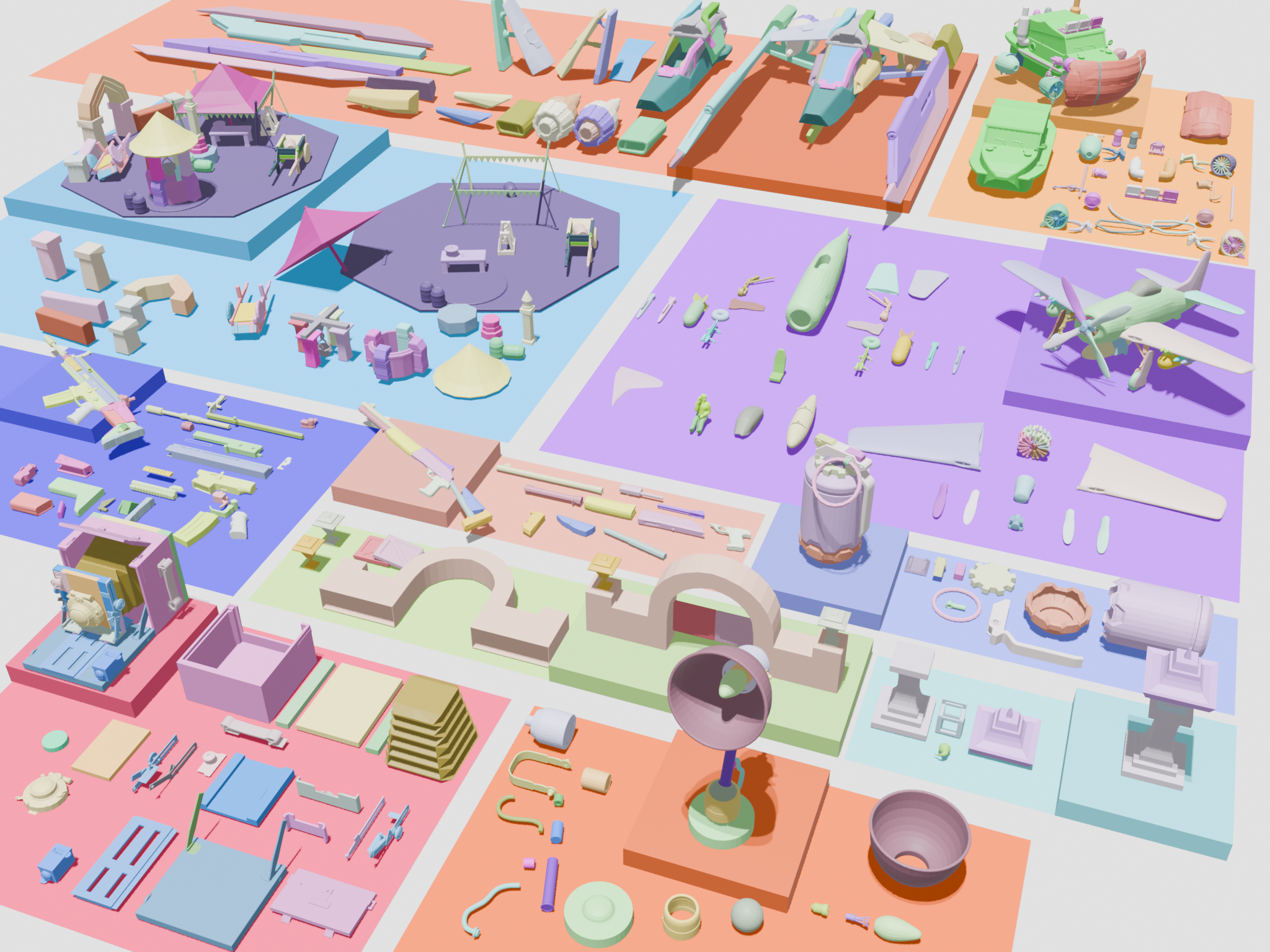}
    \caption{Part-level dataset visualization, showing individual components and the assembled model color-coded by component ID.}
    \label{fig:full_data_vis}
\end{figure}

{\bf Synthetic Data}. The Synthetic Data contains more than 125k samples. The category system design of this dataset fully considers the needs of real-world applications, ultimately encompassing 20 top-level categories, 130 mid-level subcategories, and 1,252 fine-grained classifications of product data. The breadth and depth of this category system far exceed existing real datasets, with coverage ranging from daily necessities and consumer electronics to professional industrial products.

\begin{figure}[h]
    \centering
    \includegraphics[width=0.95\linewidth]{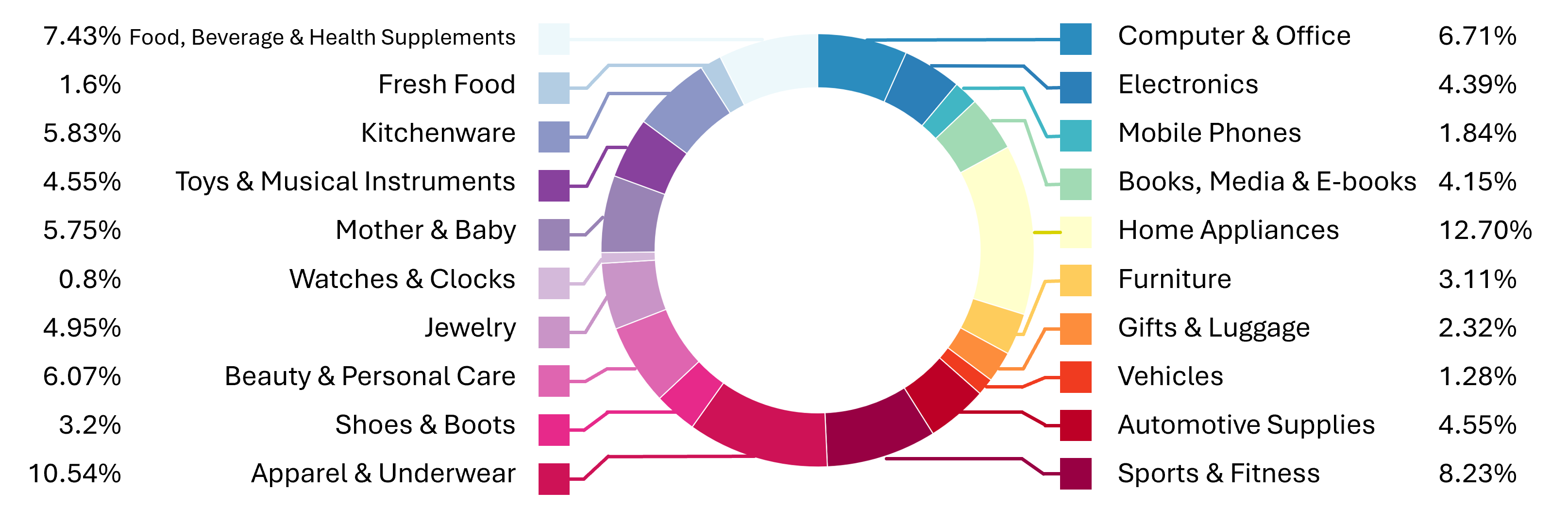}
    \caption{The Top-level Category Distribution of Synthetic Data.}
    \label{fig:component_distribution_gen_data}
\end{figure}

\begin{figure}[h]
    \centering
    \includegraphics[width=0.95\linewidth]{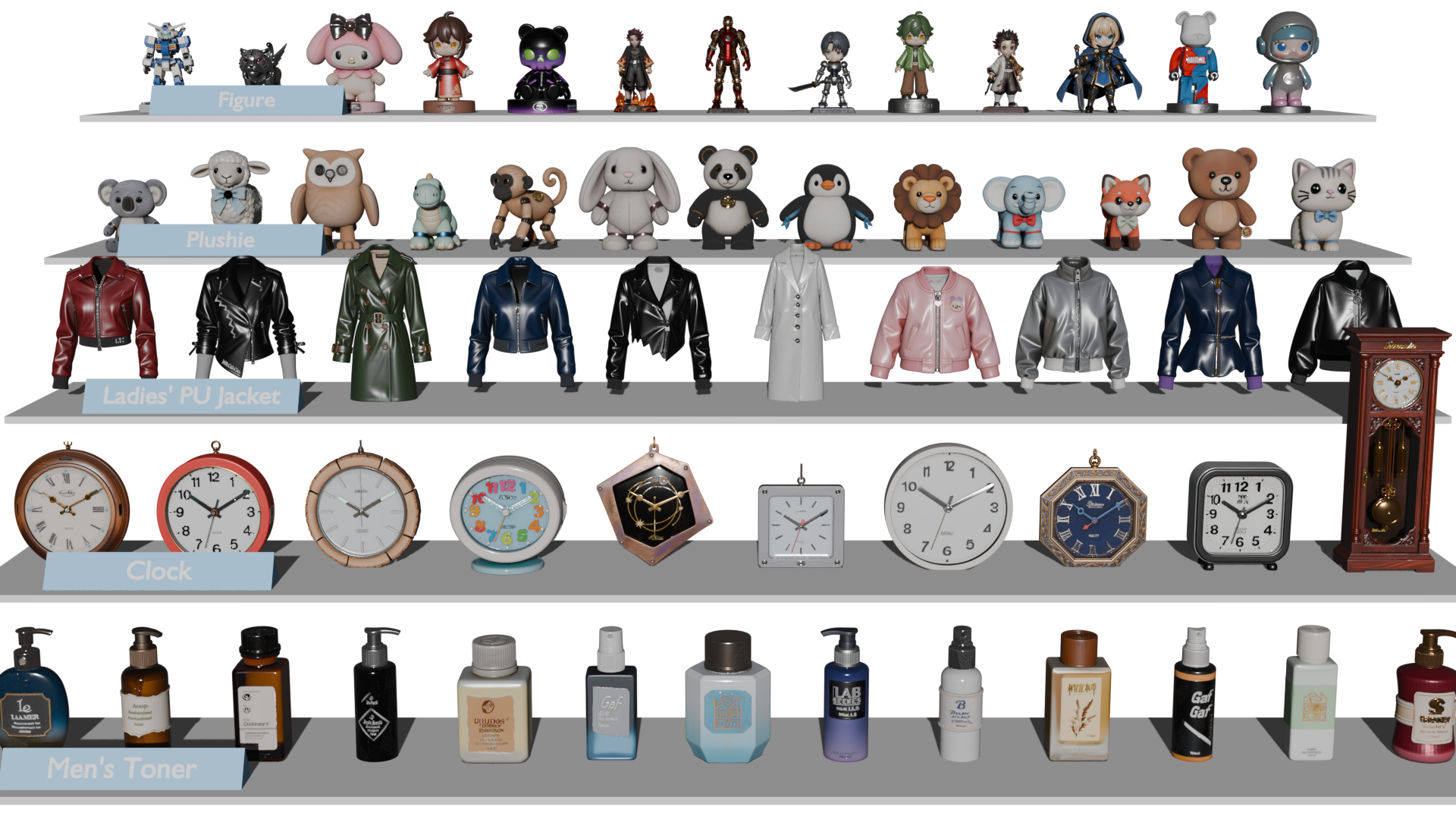}
    \caption{Synthetic dataset visualization, showing diverse samples from 5 fine-grained categories.}
    \label{fig:full_data_vis}
\end{figure}




\section{Evaluation}

\subsection{Implementation Details}
To validate the effectiveness of Full-level Data in 3D generation tasks, we use Hunyuan3D-2.1 as our baseline. While maintaining the core architectural design principles, we appropriately scale down the model to reduce training costs and train a lightweight Hunyuan3D-2.1-Small model. For evaluation, we use ULIP~\cite{xue2023ulip} and Uni3D~\cite{zhou2023uni3d} to measure the consistency between images and generated meshes.

\textbf{Model Architecture Adjustments}. Compared to the original Hunyuan3D-2.1 model, our Small model incorporates the following key architectural modifications to balance model capacity with training efficiency: (1) Channel dimension reduction: We reduce the base channel dimension from 2048 to 1536. (2) Architecture simplification: We remove the Mixture of Experts (MoE) structure and adopt a fully Dense architecture instead. After these adjustments, our Hunyuan3D-2.1-Small model contains 832M parameters.

\textbf{Progressive Training Strategy}. Drawing on the successful experience of Hunyuan3D-2.1, we employ a progressive token resolution training strategy, starting from 512 tokens and gradually increasing the token count to improve representation fidelity, ultimately reaching 4096 tokens. Detailed training configurations are provided in Table~\ref{tab:full_level_data_train}.

\begin{table}[h]
    \centering
    \begin{tabular}{c||c|c|c|c}
    \hline
         Tokens &  Batch size & Image Size & Learning rate & Traning steps \\
    \hline
         512 & 512 & 224 & 1.e-4 & 800k \\
         2048 & 256 & 224 & 5.e-5 & 400k \\
         2048 & 256 & 518 & 5.e-5 & 200k \\
         4096 & 128 & 518 & 1.e-5 & 400k \\
    \hline
    \end{tabular}
    \caption{Hunyuan3D-2.1-Small Training Strategy.}
    \label{tab:full_level_data_train}
\end{table}

\subsection{Experimental Results}

\begin{figure}[h]
    \centering
    \includegraphics[width=0.95\linewidth]{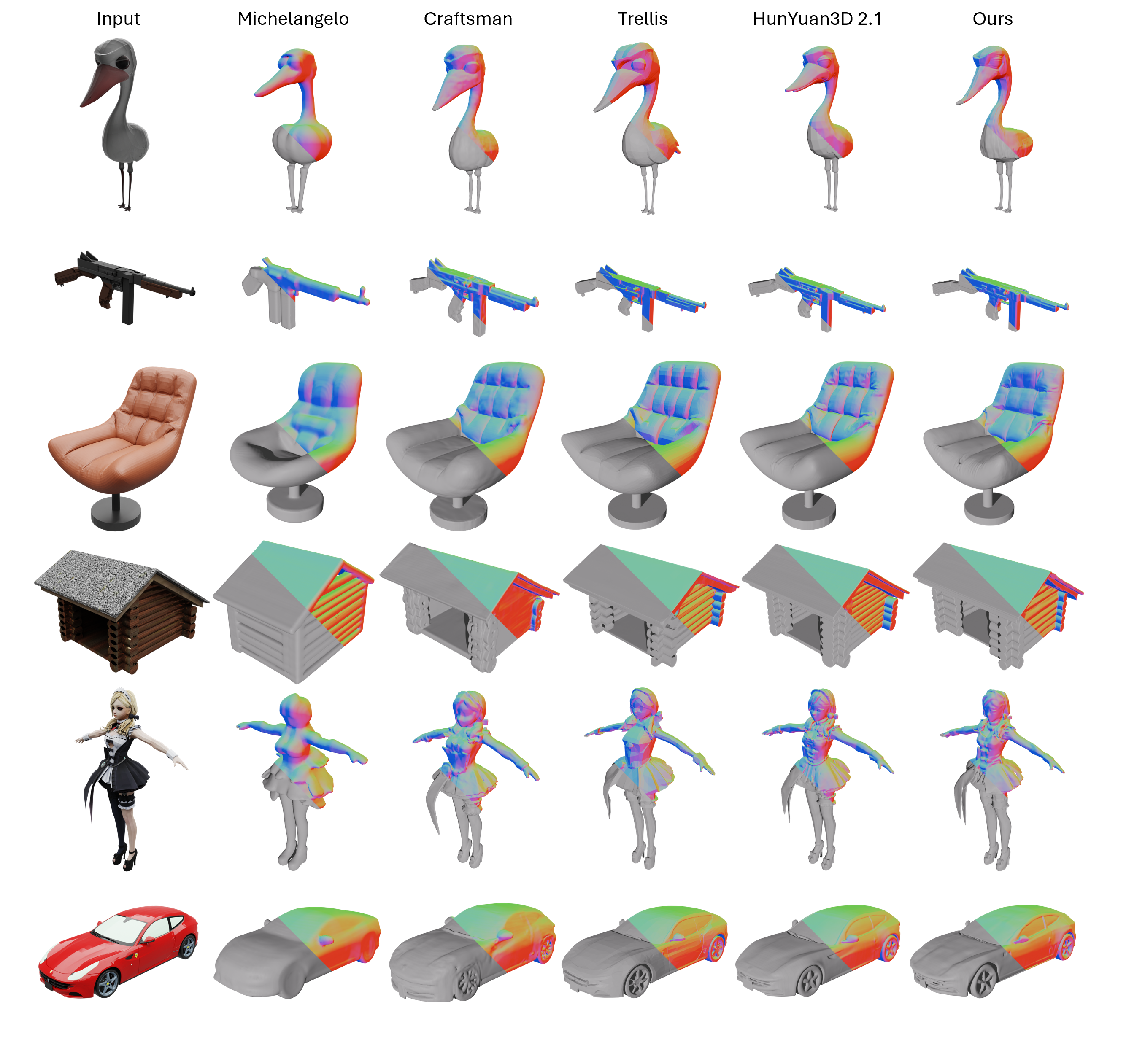}
    \caption{The qualitative comparison for image-to-3D generation on our test dataset.}
    \label{fig:full_level_data_eval}
\end{figure}

To comprehensively evaluate the effectiveness of our full-level dataset, we conducted comparative experiments with several representative state-of-the-art open-source methods, including Michelangelo~\cite{zhao2024michelangelo}, Craftsman~\cite{li2024craftsman}, Trellis~\cite{xiang2024structured}, and Hunyuan3D 2.1~\cite{hunyuan3d2025hunyuan3d}. These baseline methods have all demonstrated outstanding performance in the field of 3D generation. As shown in Table~\ref{tab:full_level_data_eval} and Figure~\ref{fig:full_level_data_eval}, despite having significantly fewer parameters than Trellis and Hunyuan3D 2.1, our model achieves comparable generation quality when trained on our open-sourced dataset, while outperforming the similarly-sized Craftsman. This experimental result fully demonstrates the high-quality characteristics of our open-sourced dataset. Meanwhile, this also indicates that data quality plays a crucial role in 3D generation tasks. The dataset we have constructed can provide the community with an efficient training resource, enabling researchers to focus more on algorithm innovation and model optimization rather than tedious data processing and preparation work.

\begin{table}[h]
    \centering
    \begin{tabular}{c|c|c|c|c}
    \hline
         Methods &  Token length & Model Size (M) & Uni3D-I $\uparrow$ & ULIP-I $\uparrow$ \\
    \hline
         Michelangelo~\cite{zhao2024michelangelo} & 257 & 105 & 0.3169 & 0.2186 \\
         CraftsMan~\cite{li2024craftsman} & 2048 & 852 & 0.3351 & 0.2264 \\
         Trellis~\cite{xiang2024structured} & 10000* & 1156 & 0.3641 & 0.2454 \\
         Hunyuan3D 2.1~\cite{hunyuan3d2025hunyuan3d} & 4096 & 1238 & 0.3636 & 0.2446 \\
         Ours  & 4096 & 832 & 0.3606 & 0.2424 \\
    \hline
    \end{tabular}
    \caption{The quantitative comparison for image-to-3D generation on our test dataset.``*'' denotes the average token length for active voxel.}
    \label{tab:full_level_data_eval}
\end{table}

\section{Conclusion}
In this work, we present HY3D-Bench, an open-source ecosystem designed to surmount the data processing bottlenecks currently constraining 3D generative models. We establish a unified foundation through three key contributions. First, we curate a high-fidelity library of 252k 3D objects derived from large-scale repositories such as Objaverse and Objaverse-XL. We employ a rigorous, multi-stage pipeline to ensure training readiness, producing essential artifacts such as watertight meshes and multi-view renderings. Second, we introduce 240k structured part-level decomposition, providing the granularity essential for advancing fine-grained perception, part-aware generation, and controllable 3D editing. Third, to mitigate real-world data distribution gaps, we develop a scalable AIGC-driven synthesis pipeline, contributing 125k synthetic assets to enrich diversity within long-tail categories. Empirical validation using the Hunyuan3D-2.1-Small model confirms the practical utility of our dataset. By democratizing access to these resources, HY3D-Bench aims to catalyze innovation across 3D perception, robotics, and digital content creation. Future efforts will focus on extending this framework to include dynamic assets and broader tasks.

\clearpage

\section{Contributors}
\large{Authors are listed \textbf{alphabetically by the first name}.} 
\definecolor{tencentblue}{RGB}{38,54,221}
\large{
\color{tencentblue}%
\begin{multicols}{2}
\raggedcolumns
Bowen Zhang\\
Chunchao Guo\\
Dongyuan Guo\\
Haolin Liu\\
Hongyu Yan\\
Huiwen Shi\\
Jiaao Yu\\
Jiachen Xu\\
Jingwei Huang\\
Kunhong Li\\
Lifu Wang\\
Linus\\
Penghao Wang\\
Qingxiang Lin\\
Ruining Tang\\
Xianghui Yang\\
Yang Li\\
Yunfei Zhao\\
Yunhan Yang\\
Zeqiang Lai\\
Zhihao Liang\\
Zibo Zhao\\
\end{multicols}}

\large{Other contributors are listed \textbf{alphabetically by the first name}.} 
\definecolor{tencentblue}{RGB}{38,54,221}
\large{
\color{tencentblue}%
\begin{multicols}{2}
\raggedcolumns
Chao Zhang\\
Edwarrd Wang\\
Hao Zhang\\
Jiaxin Lin\\
Peng He\\
Yirui Guan\\
Yonghao Tan\\
Zheng Ye\\
\end{multicols}}

\clearpage

\bibliography{colm2024_conference}

@String(CVPR= {IEEE Conf. Comput. Vis. Pattern Recog.})

@String(ICCV= {Int. Conf. Comput. Vis.})

@String(ECCV= {Eur. Conf. Comput. Vis.})

@String(TOG= {ACM Trans. Graph.})

@String(ICLR = {Int. Conf. Learn. Represent.})

@String(CVPR  = {CVPR})

@String(ICCV  = {ICCV})

@String(ECCV  = {ECCV})

@String(TOG   = {ACM TOG})

@String(ICLR  = {ICLR})

@article{goodfellow2020generative,
  title={Generative adversarial networks},
  author={Goodfellow, Ian and Pouget-Abadie, Jean and Mirza, Mehdi and Xu, Bing and Warde-Farley, David and Ozair, Sherjil and Courville, Aaron and Bengio, Yoshua},
  journal={Communications of the ACM},
  volume={63},
  number={11},
  pages={139--144},
  year={2020},
  publisher={ACM New York, NY, USA}
}

@inproceedings{luo2021surfgen,
  title={Surfgen: Adversarial 3d shape synthesis with explicit surface discriminators},
  author={Luo, Andrew and Li, Tianqin and Zhang, Wen-Hao and Lee, Tai Sing},
  booktitle={Proceedings of the IEEE/CVF international conference on computer vision},
  pages={16238--16248},
  year={2021}
}

@inproceedings{chen2022gdna,
  title={gdna: Towards generative detailed neural avatars},
  author={Chen, Xu and Jiang, Tianjian and Song, Jie and Yang, Jinlong and Black, Michael J and Geiger, Andreas and Hilliges, Otmar},
  booktitle={Proceedings of the IEEE/CVF Conference on Computer Vision and Pattern Recognition},
  pages={20427--20437},
  year={2022}
}

@article{li2021sp,
  title={SP-GAN: Sphere-guided 3D shape generation and manipulation},
  author={Li, Ruihui and Li, Xianzhi and Hui, Ka-Hei and Fu, Chi-Wing},
  journal={ACM Transactions on Graphics (TOG)},
  volume={40},
  number={4},
  pages={1--12},
  year={2021},
  publisher={ACM New York, NY, USA}
}

@inproceedings{shu20193d,
  title={3d point cloud generative adversarial network based on tree structured graph convolutions},
  author={Shu, Dong Wook and Park, Sung Woo and Kwon, Junseok},
  booktitle={Proceedings of the IEEE/CVF international conference on computer vision},
  pages={3859--3868},
  year={2019}
}

@article{wu2016learning,
  title={Learning a probabilistic latent space of object shapes via 3d generative-adversarial modeling},
  author={Wu, Jiajun and Zhang, Chengkai and Xue, Tianfan and Freeman, Bill and Tenenbaum, Josh},
  journal={Advances in neural information processing systems},
  volume={29},
  year={2016}
}

@inproceedings{nguyen2019hologan,
  title={Hologan: Unsupervised learning of 3d representations from natural images},
  author={Nguyen-Phuoc, Thu and Li, Chuan and Theis, Lucas and Richardt, Christian and Yang, Yong-Liang},
  booktitle={Proceedings of the IEEE/CVF international conference on computer vision},
  pages={7588--7597},
  year={2019}
}

@article{nguyen2020blockgan,
  title={Blockgan: Learning 3d object-aware scene representations from unlabelled images},
  author={Nguyen-Phuoc, Thu H and Richardt, Christian and Mai, Long and Yang, Yongliang and Mitra, Niloy},
  journal={Advances in neural information processing systems},
  volume={33},
  pages={6767--6778},
  year={2020}
}

@article{mildenhall2021nerf,
  title={Nerf: Representing scenes as neural radiance fields for view synthesis},
  author={Mildenhall, Ben and Srinivasan, Pratul P and Tancik, Matthew and Barron, Jonathan T and Ramamoorthi, Ravi and Ng, Ren},
  journal={Communications of the ACM},
  volume={65},
  number={1},
  pages={99--106},
  year={2021},
  publisher={ACM New York, NY, USA}
}

@article{schwarz2020graf,
  title={Graf: Generative radiance fields for 3d-aware image synthesis},
  author={Schwarz, Katja and Liao, Yiyi and Niemeyer, Michael and Geiger, Andreas},
  journal={Advances in neural information processing systems},
  volume={33},
  pages={20154--20166},
  year={2020}
}

@inproceedings{chan2021pi,
  title={pi-gan: Periodic implicit generative adversarial networks for 3d-aware image synthesis},
  author={Chan, Eric R and Monteiro, Marco and Kellnhofer, Petr and Wu, Jiajun and Wetzstein, Gordon},
  booktitle={Proceedings of the IEEE/CVF conference on computer vision and pattern recognition},
  pages={5799--5809},
  year={2021}
}

@inproceedings{chan2022efficient,
  title={Efficient geometry-aware 3d generative adversarial networks},
  author={Chan, Eric R and Lin, Connor Z and Chan, Matthew A and Nagano, Koki and Pan, Boxiao and De Mello, Shalini and Gallo, Orazio and Guibas, Leonidas J and Tremblay, Jonathan and Khamis, Sameh and others},
  booktitle={Proceedings of the IEEE/CVF conference on computer vision and pattern recognition},
  pages={16123--16133},
  year={2022}
}

@inproceedings{karras2020analyzing,
  title={Analyzing and improving the image quality of stylegan},
  author={Karras, Tero and Laine, Samuli and Aittala, Miika and Hellsten, Janne and Lehtinen, Jaakko and Aila, Timo},
  booktitle={Proceedings of the IEEE/CVF conference on computer vision and pattern recognition},
  pages={8110--8119},
  year={2020}
}

@article{poole2022dreamfusion,
  title={Dreamfusion: Text-to-3d using 2d diffusion},
  author={Poole, Ben and Jain, Ajay and Barron, Jonathan T and Mildenhall, Ben},
  journal={arXiv preprint arXiv:2209.14988},
  year={2022}
}

@article{wang2023prolificdreamer,
  title={Prolificdreamer: High-fidelity and diverse text-to-3d generation with variational score distillation},
  author={Wang, Zhengyi and Lu, Cheng and Wang, Yikai and Bao, Fan and Li, Chongxuan and Su, Hang and Zhu, Jun},
  journal={Advances in neural information processing systems},
  volume={36},
  pages={8406--8441},
  year={2023}
}

@inproceedings{liu2023zero,
  title={Zero-1-to-3: Zero-shot one image to 3d object},
  author={Liu, Ruoshi and Wu, Rundi and Van Hoorick, Basile and Tokmakov, Pavel and Zakharov, Sergey and Vondrick, Carl},
  booktitle={Proceedings of the IEEE/CVF international conference on computer vision},
  pages={9298--9309},
  year={2023}
}

@article{EnVision2023luciddreamer,
  title={Luciddreamer: Towards high-fidelity text-to-3d generation via interval score matching},
  author={Liang, Yixun and Yang, Xin and Lin, Jiantao and Li, Haodong and Xu, Xiaogang and Chen, Yingcong},
  journal={arXiv preprint arXiv:2311.11284},
  year={2023}
}

@InProceedings{fantasia3d,
author    = {Chen, Rui and Chen, Yongwei and Jiao, Ningxin and Jia, Kui},
title     = {Fantasia3D: Disentangling Geometry and Appearance for High-quality Text-to-3D Content Creation},
booktitle = {Proceedings of the IEEE/CVF International Conference on Computer Vision (ICCV)},
month     = {October},
year      = {2023}
}

@article{sweetdreamer,
author    = {Weiyu Li and Rui Chen and Xuelin Chen and Ping Tan},
title     = {SweetDreamer: Aligning Geometric Priors in 2D Diffusion for Consistent Text-to-3D},
journal   = ICLR,
year      = {2024},
}

@inproceedings{yi2024gaussiandreamer,
  title={Gaussiandreamer: Fast generation from text to 3d gaussians by bridging 2d and 3d diffusion models},
  author={Yi, Taoran and Fang, Jiemin and Wang, Junjie and Wu, Guanjun and Xie, Lingxi and Zhang, Xiaopeng and Liu, Wenyu and Tian, Qi and Wang, Xinggang},
  booktitle={Proceedings of the IEEE/CVF Conference on Computer Vision and Pattern Recognition},
  pages={6796--6807},
  year={2024}
}

@article{zhang20233dshape2vecset,
  title={3dshape2vecset: A 3d shape representation for neural fields and generative diffusion models},
  author={Zhang, Biao and Tang, Jiapeng and Niessner, Matthias and Wonka, Peter},
  journal={ACM Transactions On Graphics (TOG)},
  volume={42},
  number={4},
  pages={1--16},
  year={2023},
  publisher={ACM New York, NY, USA}
}

@inproceedings{tang2024lgm,
  title={Lgm: Large multi-view gaussian model for high-resolution 3d content creation},
  author={Tang, Jiaxiang and Chen, Zhaoxi and Chen, Xiaokang and Wang, Tengfei and Zeng, Gang and Liu, Ziwei},
  booktitle={European Conference on Computer Vision},
  pages={1--18},
  year={2024},
  organization={Springer}
}

@article{li2023instant3d,
  title={Instant3d: Fast text-to-3d with sparse-view generation and large reconstruction model},
  author={Li, Jiahao and Tan, Hao and Zhang, Kai and Xu, Zexiang and Luan, Fujun and Xu, Yinghao and Hong, Yicong and Sunkavalli, Kalyan and Shakhnarovich, Greg and Bi, Sai},
  journal={arXiv preprint arXiv:2311.06214},
  year={2023}
}

@article{hong2023lrm,
  title={Lrm: Large reconstruction model for single image to 3d},
  author={Hong, Yicong and Zhang, Kai and Gu, Jiuxiang and Bi, Sai and Zhou, Yang and Liu, Difan and Liu, Feng and Sunkavalli, Kalyan and Bui, Trung and Tan, Hao},
  journal={arXiv preprint arXiv:2311.04400},
  year={2023}
}

@article{li2025triposg,
  title={TripoSG: High-Fidelity 3D Shape Synthesis using Large-Scale Rectified Flow Models},
  author={Li, Yangguang and Zou, Zi-Xin and Liu, Zexiang and Wang, Dehu and Liang, Yuan and Yu, Zhipeng and Liu, Xingchao and Guo, Yuan-Chen and Liang, Ding and Ouyang, Wanli and others},
  journal={arXiv preprint arXiv:2502.06608},
  year={2025}
}

@article{huang2025spar3d,
  title={SPAR3D: Stable Point-Aware Reconstruction of 3D Objects from Single Images},
  author={Huang, Zixuan and Boss, Mark and Vasishta, Aaryaman and Rehg, James M and Jampani, Varun},
  journal={arXiv preprint arXiv:2501.04689},
  year={2025}
}

@misc{li2024craftsman,
title         = {CraftsMan: High-fidelity Mesh Generation with 3D Native Generation and Interactive Geometry Refiner}, 
author        = {Weiyu Li and Jiarui Liu and Hongyu Yan and Rui Chen and Yixun Liang and Xuelin Chen and Ping Tan and Xiaoxiao Long},
year          = {2024},
archivePrefix = {arXiv preprint arXiv:2405.14979},
primaryClass  = {cs.CG}
}

@article{zhang2024clay,
  title={CLAY: A Controllable Large-scale Generative Model for Creating High-quality 3D Assets},
  author={Zhang, Longwen and Wang, Ziyu and Zhang, Qixuan and Qiu, Qiwei and Pang, Anqi and Jiang, Haoran and Yang, Wei and Xu, Lan and Yu, Jingyi},
  journal={ACM Transactions on Graphics (TOG)},
  volume={43},
  number={4},
  pages={1--20},
  year={2024},
  publisher={ACM New York, NY, USA}
}

@article{zhao2024michelangelo,
  title={Michelangelo: Conditional 3d shape generation based on shape-image-text aligned latent representation},
  author={Zhao, Zibo and Liu, Wen and Chen, Xin and Zeng, Xianfang and Wang, Rui and Cheng, Pei and Fu, Bin and Chen, Tao and Yu, Gang and Gao, Shenghua},
  journal={Advances in Neural Information Processing Systems},
  volume={36},
  year={2024}
}

@article{xiang2024structured,
    title   = {Structured 3D Latents for Scalable and Versatile 3D Generation},
    author  = {Xiang, Jianfeng and Lv, Zelong and Xu, Sicheng and Deng, Yu and Wang, Ruicheng and 
               Zhang, Bowen and Chen, Dong and Tong, Xin and Yang, Jiaolong},
    journal = {arXiv preprint arXiv:2412.01506},
    year    = {2024}
}

@article{wu2024direct3d,
  title={Direct3D: Scalable Image-to-3D Generation via 3D Latent Diffusion Transformer},
  author={Wu, Shuang and Lin, Youtian and Zhang, Feihu and Zeng, Yifei and Xu, Jingxi and Torr, Philip and Cao, Xun and Yao, Yao},
  journal={arXiv preprint arXiv:2405.14832},
  year={2024}
}

@inproceedings{xue2023ulip,
  title={Ulip: Learning a unified representation of language, images, and point clouds for 3d understanding},
  author={Xue, Le and Gao, Mingfei and Xing, Chen and Mart{\'\i}n-Mart{\'\i}n, Roberto and Wu, Jiajun and Xiong, Caiming and Xu, Ran and Niebles, Juan Carlos and Savarese, Silvio},
  booktitle={Proceedings of the IEEE/CVF conference on computer vision and pattern recognition},
  pages={1179--1189},
  year={2023}
}

@article{zhou2023uni3d,
  title={Uni3d: Exploring unified 3d representation at scale},
  author={Zhou, Junsheng and Wang, Jinsheng and Ma, Baorui and Liu, Yu-Shen and Huang, Tiejun and Wang, Xinlong},
  journal={arXiv preprint arXiv:2310.06773},
  year={2023}
}

@article{wang2023pf,
  title={Pf-lrm: Pose-free large reconstruction model for joint pose and shape prediction},
  author={Wang, Peng and Tan, Hao and Bi, Sai and Xu, Yinghao and Luan, Fujun and Sunkavalli, Kalyan and Wang, Wenping and Xu, Zexiang and Zhang, Kai},
  journal={arXiv preprint arXiv:2311.12024},
  year={2023}
}

@inproceedings{long2024wonder3d,
  title={Wonder3d: Single image to 3d using cross-domain diffusion},
  author={Long, Xiaoxiao and Guo, Yuan-Chen and Lin, Cheng and Liu, Yuan and Dou, Zhiyang and Liu, Lingjie and Ma, Yuexin and Zhang, Song-Hai and Habermann, Marc and Theobalt, Christian and others},
  booktitle={Proceedings of the IEEE/CVF conference on computer vision and pattern recognition},
  pages={9970--9980},
  year={2024}
}

@article{ye2025nano3d,
  title={NANO3D: A Training-Free Approach for Efficient 3D Editing Without Masks},
  author={Ye, Junliang and Xie, Shenghao and Zhao, Ruowen and Wang, Zhengyi and Yan, Hongyu and Zu, Wenqiang and Ma, Lei and Zhu, Jun},
  journal={arXiv preprint arXiv:2510.15019},
  year={2025}
}

@inproceedings{park2019deepsdf,
  title={Deepsdf: Learning continuous signed distance functions for shape representation},
  author={Park, Jeong Joon and Florence, Peter and Straub, Julian and Newcombe, Richard and Lovegrove, Steven},
  booktitle=CVPR,
  pages={165--174},
  year={2019}
}

@inproceedings{luo2021diffusion,
  author = {Luo, Shitong and Hu, Wei},
  title = {Diffusion Probabilistic Models for 3D Point Cloud Generation},
  booktitle = {Proceedings of the IEEE/CVF Conference on Computer Vision and Pattern Recognition (CVPR)},
  month = {June},
  year = {2021}
}

@inproceedings{chen2019learning,
  title={Learning implicit fields for generative shape modeling},
  author={Chen, Zhiqin and Zhang, Hao},
  booktitle=CVPR,
  pages={5939--5948},
  year={2019}
}

@inproceedings{Liu2023MeshDiffusion,
    title={MeshDiffusion: Score-based Generative 3D Mesh Modeling},
    author={Zhen Liu and Yao Feng and Michael J. Black and Derek Nowrouzezahrai and Liam Paull and Weiyang Liu},
    booktitle={International Conference on Learning Representations},
    year={2023},
    url={https://openreview.net/forum?id=0cpM2ApF9p6}
}

@inproceedings{nash2020polygen,
  title={Polygen: An autoregressive generative model of 3d meshes},
  author={Nash, Charlie and Ganin, Yaroslav and Eslami, SM Ali and Battaglia, Peter},
  booktitle={International conference on machine learning},
  pages={7220--7229},
  year={2020},
  organization={PMLR}
}

@inproceedings{zhou20213d,
  title={3d shape generation and completion through point-voxel diffusion},
  author={Zhou, Linqi and Du, Yilun and Wu, Jiajun},
  booktitle={Proceedings of the IEEE/CVF international conference on computer vision},
  pages={5826--5835},
  year={2021}
}

@article{pointflow,
 title={PointFlow: 3D Point Cloud Generation with Continuous Normalizing Flows},
 author={Yang, Guandao and Huang, Xun and Hao, Zekun and Liu, Ming-Yu and Belongie, Serge and Hariharan, Bharath},
 journal={arXiv},
 year={2019}
}

@article{wu2025direct3d,
  title={Direct3d-s2: Gigascale 3d generation made easy with spatial sparse attention},
  author={Wu, Shuang and Lin, Youtian and Zhang, Feihu and Zeng, Yifei and Yang, Yikang and Bao, Yajie and Qian, Jiachen and Zhu, Siyu and Cao, Xun and Torr, Philip and others},
  journal={arXiv preprint arXiv:2505.17412},
  year={2025}
}

@article{he2025sparseflex,
  title={Sparseflex: High-resolution and arbitrary-topology 3d shape modeling},
  author={He, Xianglong and Zou, Zi-Xin and Chen, Chia-Hao and Guo, Yuan-Chen and Liang, Ding and Yuan, Chun and Ouyang, Wanli and Cao, Yan-Pei and Li, Yangguang},
  journal={arXiv preprint arXiv:2503.21732},
  year={2025}
}

@article{hunyuan3d2025omni,
  title={Hunyuan3d-omni: A unified framework for controllable generation of 3d assets},
  author={Hunyuan3D, Team and Zhang, Bowen and Guo, Chunchao and Liu, Haolin and Yan, Hongyu and Shi, Huiwen and Huang, Jingwei and Yu, Junlin and Li, Kunhong and Wang, Penghao and others},
  journal={arXiv preprint arXiv:2509.21245},
  year={2025}
}

@article{yan2025posemaster,
  title={PoseMaster: Generating 3D Characters in Arbitrary Poses from a Single Image},
  author={Yan, Hongyu and Luo, Kunming and Li, Weiyu and Liang, Yixun and Li, Shengming and Huang, Jingwei and Guo, Chunchao and Tan, Ping},
  journal={arXiv preprint arXiv:2506.21076},
  year={2025}
}

@article{dong2025crossgen,
  title={CrossGen: Learning and Generating Cross Fields for Quad Meshing},
  author={Dong, Qiujie and Wang, Jiepeng and Xu, Rui and Lin, Cheng and Liu, Yuan and Xin, Shiqing and Zhong, Zichun and Li, Xin and Tu, Changhe and Komura, Taku and others},
  journal={arXiv preprint arXiv:2506.07020},
  year={2025}
}

@article{hunyuan3d2025hunyuan3d,
  title={Hunyuan3D 2.1: From Images to High-Fidelity 3D Assets with Production-Ready PBR Material},
  author={Hunyuan3D, Team and Yang, Shuhui and Yang, Mingxin and Feng, Yifei and Huang, Xin and Zhang, Sheng and He, Zebin and Luo, Di and Liu, Haolin and Zhao, Yunfei and others},
  journal={arXiv preprint arXiv:2506.15442},
  year={2025}
}

@article{liu2023syncdreamer,
  title={Syncdreamer: Generating multiview-consistent images from a single-view image},
  author={Liu, Yuan and Lin, Cheng and Zeng, Zijiao and Long, Xiaoxiao and Liu, Lingjie and Komura, Taku and Wang, Wenping},
  journal={arXiv preprint arXiv:2309.03453},
  year={2023}
}

@article{li2024era3d,
  title={Era3d: High-resolution multiview diffusion using efficient row-wise attention},
  author={Li, Peng and Liu, Yuan and Long, Xiaoxiao and Zhang, Feihu and Lin, Cheng and Li, Mengfei and Qi, Xingqun and Zhang, Shanghang and Xue, Wei and Luo, Wenhan and others},
  journal={Advances in Neural Information Processing Systems},
  volume={37},
  pages={55975--56000},
  year={2024}
}

@article{liu2023one,
  title={One-2-3-45: Any single image to 3d mesh in 45 seconds without per-shape optimization},
  author={Liu, Minghua and Xu, Chao and Jin, Haian and Chen, Linghao and Varma T, Mukund and Xu, Zexiang and Su, Hao},
  journal={Advances in Neural Information Processing Systems},
  volume={36},
  pages={22226--22246},
  year={2023}
}

@article{kerbl20233d,
  title={3D Gaussian splatting for real-time radiance field rendering.},
  author={Kerbl, Bernhard and Kopanas, Georgios and Leimk{\"u}hler, Thomas and Drettakis, George},
  journal={ACM Trans. Graph.},
  volume={42},
  number={4},
  pages={139--1},
  year={2023}
}

@inproceedings{ren2024xcube,
  title={Xcube: Large-scale 3d generative modeling using sparse voxel hierarchies},
  author={Ren, Xuanchi and Huang, Jiahui and Zeng, Xiaohui and Museth, Ken and Fidler, Sanja and Williams, Francis},
  booktitle={Proceedings of the IEEE/CVF Conference on Computer Vision and Pattern Recognition},
  pages={4209--4219},
  year={2024}
}

@article{chang2015shapenet,
  title={Shapenet: An information-rich 3d model repository},
  author={Chang, Angel X and Funkhouser, Thomas and Guibas, Leonidas and Hanrahan, Pat and Huang, Qixing and Li, Zimo and Savarese, Silvio and Savva, Manolis and Song, Shuran and Su, Hao and others},
  journal={arXiv preprint arXiv:1512.03012},
  year={2015}
}

@article{zhou2016thingi10k,
  title={Thingi10k: A dataset of 10,000 3d-printing models},
  author={Zhou, Qingnan and Jacobson, Alec},
  journal={arXiv preprint arXiv:1605.04797},
  year={2016}
}

@inproceedings{fu20213d,
  title={3d-front: 3d furnished rooms with layouts and semantics},
  author={Fu, Huan and Cai, Bowen and Gao, Lin and Zhang, Ling-Xiao and Wang, Jiaming and Li, Cao and Zeng, Qixun and Sun, Chengyue and Jia, Rongfei and Zhao, Binqiang and others},
  booktitle={Proceedings of the IEEE/CVF International Conference on Computer Vision},
  pages={10933--10942},
  year={2021}
}

@article{collins2022abo,
  title={ABO: Dataset and Benchmarks for Real-World 3D Object Understanding},
  author={Collins, Jasmine and Goel, Shubham and Deng, Kenan and Luthra, Achleshwar and
          Xu, Leon and Gundogdu, Erhan and Zhang, Xi and Yago Vicente, Tomas F and
          Dideriksen, Thomas and Arora, Himanshu and Guillaumin, Matthieu and
          Malik, Jitendra},
  journal={CVPR},
  year={2022}
}

@article{objaverse,
  title={Objaverse: A Universe of Annotated 3D Objects},
  author={Matt Deitke and Dustin Schwenk and Jordi Salvador and Luca Weihs and
          Oscar Michel and Eli VanderBilt and Ludwig Schmidt and
          Kiana Ehsani and Aniruddha Kembhavi and Ali Farhadi},
  journal={arXiv preprint arXiv:2212.08051},
  year={2022}
}

@article{objaverseXL,
  title={Objaverse-XL: A Universe of 10M+ 3D Objects},
  author={Matt Deitke and Ruoshi Liu and Matthew Wallingford and Huong Ngo and
          Oscar Michel and Aditya Kusupati and Alan Fan and Christian Laforte and
          Vikram Voleti and Samir Yitzhak Gadre and Eli VanderBilt and
          Aniruddha Kembhavi and Carl Vondrick and Georgia Gkioxari and
          Kiana Ehsani and Ludwig Schmidt and Ali Farhadi},
  journal={arXiv preprint arXiv:2307.05663},
  year={2023}
}

@inproceedings{shimvdream,
  title={MVDream: Multi-view Diffusion for 3D Generation},
  author={Shi, Yichun and Wang, Peng and Ye, Jianglong and Mai, Long and Li, Kejie and Yang, Xiao},
  booktitle={The Twelfth International Conference on Learning Representations},
  year={2023}
}

@article{xu2024instantmesh,
  title={Instantmesh: Efficient 3d mesh generation from a single image with sparse-view large reconstruction models},
  author={Xu, Jiale and Cheng, Weihao and Gao, Yiming and Wang, Xintao and Gao, Shenghua and Shan, Ying},
  journal={arXiv preprint arXiv:2404.07191},
  year={2024}
}

@inproceedings{wu2023omniobject3d,
  title={Omniobject3d: Large-vocabulary 3d object dataset for realistic perception, reconstruction and generation},
  author={Wu, Tong and Zhang, Jiarui and Fu, Xiao and Wang, Yuxin and Ren, Jiawei and Pan, Liang and Wu, Wayne and Yang, Lei and Wang, Jiaqi and Qian, Chen and others},
  booktitle={Proceedings of the IEEE/CVF Conference on Computer Vision and Pattern Recognition},
  pages={803--814},
  year={2023}
}

@inproceedings{downs2022google,
  title={Google scanned objects: A high-quality dataset of 3d scanned household items},
  author={Downs, Laura and Francis, Anthony and Koenig, Nate and Kinman, Brandon and Hickman, Ryan and Reymann, Krista and McHugh, Thomas B and Vanhoucke, Vincent},
  booktitle={2022 International Conference on Robotics and Automation (ICRA)},
  pages={2553--2560},
  year={2022},
  organization={IEEE}
}

@article{tang2023dreamgaussian,
  title={Dreamgaussian: Generative gaussian splatting for efficient 3d content creation},
  author={Tang, Jiaxiang and Ren, Jiawei and Zhou, Hang and Liu, Ziwei and Zeng, Gang},
  journal={arXiv preprint arXiv:2309.16653},
  year={2023}
}

@inproceedings{lin2023magic3d,
  title={Magic3D: High-Resolution Text-to-3D Content Creation},
  author={Lin, Chen-Hsuan and Gao, Jun and Tang, Luming and Takikawa, Towaki and Zeng, Xiaohui and Huang, Xun and Kreis, Karsten and Fidler, Sanja and Liu, Ming-Yu and Lin, Tsung-Yi},
  booktitle=CVPR,
  year={2023}
}

@article{yan2025x,
  title={X-Part: high fidelity and structure coherent shape decomposition},
  author={Yan, Xinhao and Xu, Jiachen and Li, Yang and Ma, Changfeng and Yang, Yunhan and Wang, Chunshi and Zhao, Zibo and Lai, Zeqiang and Zhao, Yunfei and Chen, Zhuo and others},
  journal={arXiv preprint arXiv:2509.08643},
  year={2025}
}

@inproceedings{liu2023partslip,
  title={Partslip: Low-shot part segmentation for 3d point clouds via pretrained image-language models},
  author={Liu, Minghua and Zhu, Yinhao and Cai, Hong and Han, Shizhong and Ling, Zhan and Porikli, Fatih and Su, Hao},
  booktitle={CVPR},
  year={2023}
}

@inproceedings{kim2024partstad,
  title={PartSTAD: 2D-to-3D Part Segmentation Task Adaptation},
  author={Kim, Hyunjin and Sung, Minhyuk},
  booktitle={ECCV},
  year={2024}
}

@inproceedings{zhong2024meshsegmenter,
  title={MeshSegmenter: Zero-Shot Mesh Semantic Segmentation via Texture Synthesis},
  author={Zhong, Ziming and Xu, Yanyu and Li, Jing and Xu, Jiale and Li, Zhengxin and Yu, Chaohui and Gao, Shenghua},
  booktitle={ECCV},
  year={2024},
}

@inproceedings{abdelreheem2023satr,
  title={Satr: Zero-shot semantic segmentation of 3d shapes},
  author={Abdelreheem, Ahmed and Skorokhodov, Ivan and Ovsjanikov, Maks and Wonka, Peter},
  booktitle={ICCV},
  year={2023}
}

@article{tang2024segment,
  title={Segment Any Mesh: Zero-shot Mesh Part Segmentation via Lifting Segment Anything 2 to 3D},
  author={Tang, George and Zhao, William and Ford, Logan and Benhaim, David and Zhang, Paul},
  journal={arXiv:2408.13679},
  year={2024}
}

@inproceedings{thai20243x2,
  title={3x2: 3D Object Part Segmentation by 2D Semantic Correspondences},
  author={Thai, Anh and Wang, Weiyao and Tang, Hao and Stojanov, Stefan and Feiszli, Matt and Rehg, James M},
  booktitle={ECCV},
  year={2024}
}

@article{xue2023zerops,
  title={ZeroPS: High-quality Cross-modal Knowledge Transfer for Zero-Shot 3D Part Segmentation},
  author={Xue, Yuheng and Chen, Nenglun and Liu, Jun and Sun, Wenyun},
  journal={arXiv:2311.14262},
  year={2023}
}

@article{yang2024sampart3d,
  title={Sampart3d: Segment any part in 3d objects},
  author={Yang, Yunhan and Huang, Yukun and Guo, Yuan-Chen and Lu, Liangjun and Wu, Xiaoyang and Lam, Edmund Y and Cao, Yan-Pei and Liu, Xihui},
  journal={arXiv preprint arXiv:2411.07184},
  year={2024}
}

@article{nichol2022point,
  title={Point-e: A system for generating 3d point clouds from complex prompts},
  author={Nichol, Alex and Jun, Heewoo and Dhariwal, Prafulla and Mishkin, Pamela and Chen, Mark},
  journal={arXiv preprint arXiv:2212.08751},
  year={2022}
}

@article{jun2023shap,
  title={Shap-e: Generating conditional 3d implicit functions},
  author={Jun, Heewoo and Nichol, Alex},
  journal={arXiv preprint arXiv:2305.02463},
  year={2023}
}

@inproceedings{liu2024part123,
  title={Part123: part-aware 3d reconstruction from a single-view image},
  author={Liu, Anran and Lin, Cheng and Liu, Yuan and Long, Xiaoxiao and Dou, Zhiyang and Guo, Hao-Xiang and Luo, Ping and Wang, Wenping},
  booktitle={ACM SIGGRAPH},
  year={2024}
}

@article{deng2025geosam2,
  title={GeoSAM2: Unleashing the Power of SAM2 for 3D Part Segmentation},
  author={Deng, Ken and Yang, Yunhan and Sun, Jingxiang and Liu, Xihui and Liu, Yebin and Liang, Ding and Cao, Yan-Pei},
  journal={arXiv preprint arXiv:2508.14036},
  year={2025}
}

@article{yang2025holopart,
  title={HoloPart: Generative 3D Part Amodal Segmentation},
  author={Yang, Yunhan and Guo, Yuan-Chen and Huang, Yukun and Zou, Zi-Xin and Yu, Zhipeng and Li, Yangguang and Cao, Yan-Pei and Liu, Xihui},
  journal={arXiv preprint arXiv:2504.07943},
  year={2025}
}

@article{ma2025p3,
  title={P3-sam: Native 3d part segmentation},
  author={Ma, Changfeng and Li, Yang and Yan, Xinhao and Xu, Jiachen and Yang, Yunhan and Wang, Chunshi and Zhao, Zibo and Guo, Yanwen and Chen, Zhuo and Guo, Chunchao},
  journal={arXiv preprint arXiv:2509.06784},
  year={2025}
}

@article{zhou2024point,
  title={Point-sam: Promptable 3d segmentation model for point clouds},
  author={Zhou, Yuchen and Gu, Jiayuan and Chiang, Tung Yen and Xiang, Fanbo and Su, Hao},
  journal={arXiv preprint arXiv:2406.17741},
  year={2024}
}

@article{fischer2024sama,
  title={SAMa: Material-aware 3D selection and segmentation},
  author={Fischer, Michael and Georgiev, Iliyan and Groueix, Thibault and Kim, Vladimir G and Ritschel, Tobias and Deschaintre, Valentin},
  journal={arXiv preprint arXiv:2411.19322},
  year={2024}
}

@inproceedings{ma2025find,
  title={Find any part in 3d},
  author={Ma, Ziqi and Yue, Yisong and Gkioxari, Georgia},
  booktitle={Proceedings of the IEEE/CVF International Conference on Computer Vision},
  pages={7818--7827},
  year={2025}
}

@inproceedings{liu2025partfield,
  title={Partfield: Learning 3d feature fields for part segmentation and beyond},
  author={Liu, Minghua and Uy, Mikaela Angelina and Xiang, Donglai and Su, Hao and Fidler, Sanja and Sharp, Nicholas and Gao, Jun},
  booktitle={Proceedings of the IEEE/CVF International Conference on Computer Vision},
  pages={9704--9715},
  year={2025}
}

@article{lin2025partcrafter,
  title={PartCrafter: Structured 3D Mesh Generation via Compositional Latent Diffusion Transformers},
  author={Lin, Yuchen and Lin, Chenguo and Pan, Panwang and Yan, Honglei and Feng, Yiqiang and Mu, Yadong and Fragkiadaki, Katerina},
  journal={arXiv preprint arXiv:2506.05573},
  year={2025}
}

@article{tang2025efficient,
  title={Efficient Part-level 3D Object Generation via Dual Volume Packing},
  author={Tang, Jiaxiang and Lu, Ruijie and Li, Zhaoshuo and Hao, Zekun and Li, Xuan and Wei, Fangyin and Song, Shuran and Zeng, Gang and Liu, Ming-Yu and Lin, Tsung-Yi},
  journal={arXiv preprint arXiv:2506.09980},
  year={2025}
}

@inproceedings{yang2025omnipart,
  title={Omnipart: Part-aware 3d generation with semantic decoupling and structural cohesion},
  author={Yang, Yunhan and Zhou, Yufan and Guo, Yuan-Chen and Zou, Zi-Xin and Huang, Yukun and Liu, Ying-Tian and Xu, Hao and Liang, Ding and Cao, Yan-Pei and Liu, Xihui},
  booktitle={Proceedings of the SIGGRAPH Asia 2025 Conference Papers},
  pages={1--12},
  year={2025}
}

@article{ding2025fullpart,
  title={FullPart: Generating each 3D Part at Full Resolution},
  author={Ding, Lihe and Dong, Shaocong and Li, Yaokun and Gao, Chenjian and Chen, Xiao and Han, Rui and Kuang, Yihao and Zhang, Hong and Huang, Bo and Huang, Zhanpeng and others},
  journal={arXiv preprint arXiv:2510.26140},
  year={2025}
}

@article{he2025unipart,
  title={UniPart: Part-Level 3D Generation with Unified 3D Geom-Seg Latents},
  author={He, Xufan and Wu, Yushuang and Guo, Xiaoyang and Ye, Chongjie and Zhou, Jiaqing and Hu, Tianlei and Han, Xiaoguang and Du, Dong},
  journal={arXiv preprint arXiv:2512.09435},
  year={2025}
}

@inproceedings{dong2025one,
  title={From one to more: Contextual part latents for 3d generation},
  author={Dong, Shaocong and Ding, Lihe and Chen, Xiao and Li, Yaokun and Wang, Yuxin and Wang, Yucheng and Wang, Qi and Kim, Jaehyeok and Gao, Chenjian and Huang, Zhanpeng and others},
  booktitle={Proceedings of the IEEE/CVF International Conference on Computer Vision},
  pages={8230--8240},
  year={2025}
}

@article{li2025moca,
  title={MoCA: Mixture-of-Components Attention for Scalable Compositional 3D Generation},
  author={Li, Zhiqi and Li, Wenhuan and Wang, Tengfei and Wang, Zhenwei and Wu, Junta and Wang, Haoyuan and Yang, Yunhan and Huang, Zehuan and Li, Yang and Liu, Peidong and others},
  journal={arXiv preprint arXiv:2512.07628},
  year={2025}
}

@article{zhu2025partsam,
  title={PartSAM: A Scalable Promptable Part Segmentation Model Trained on Native 3D Data},
  author={Zhu, Zhe and Wan, Le and Xu, Rui and Zhang, Yiheng and Chen, Honghua and Dou, Zhiyang and Lin, Cheng and Liu, Yuan and Wei, Mingqiang},
  journal={arXiv preprint arXiv:2509.21965},
  year={2025}
}

@article{paul2025name,
  title={Name That Part: 3D Part Segmentation and Naming},
  author={Paul, Soumava and Kaushik, Prakhar and Vaidya, Ankit and Bhattad, Anand and Yuille, Alan},
  journal={arXiv preprint arXiv:2512.18003},
  year={2025}
}

@article{li2025auto,
  title={Auto-Regressive Surface Cutting},
  author={Li, Yang and Cheung, Victor and Liu, Xinhai and Chen, Yuguang and Luo, Zhongjin and Lei, Biwen and Weng, Haohan and Zhao, Zibo and Huang, Jingwei and Chen, Zhuo and others},
  journal={arXiv preprint arXiv:2506.18017},
  year={2025}
}

@article{yang2025partdiffuser,
  title={PartDiffuser: Part-wise 3D Mesh Generation via Discrete Diffusion},
  author={Yang, Yichen and Li, Hong and Zhu, Haodong and Yang, Linin and Lei, Guojun and Xu, Sheng and Zhang, Baochang},
  journal={arXiv preprint arXiv:2511.18801},
  year={2025}
}

@inproceedings{mo2019partnet,
  title={Partnet: A large-scale benchmark for fine-grained and hierarchical part-level 3d object understanding},
  author={Mo, Kaichun and Zhu, Shilin and Chang, Angel X and Yi, Li and Tripathi, Subarna and Guibas, Leonidas J and Su, Hao},
  booktitle={Proceedings of the IEEE/CVF conference on computer vision and pattern recognition},
  pages={909--918},
  year={2019}
}

@article{wang2025partnext,
  title={PartNeXt: A Next-Generation Dataset for Fine-Grained and Hierarchical 3D Part Understanding},
  author={Wang, Penghao and He, Yiyang and Lv, Xin and Zhou, Yukai and Xu, Lan and Yu, Jingyi and Gu, Jiayuan},
  journal={arXiv preprint arXiv:2510.20155},
  year={2025}
}

@inproceedings{geng2023gapartnet,
  title={Gapartnet: Cross-category domain-generalizable object perception and manipulation via generalizable and actionable parts},
  author={Geng, Haoran and Xu, Helin and Zhao, Chengyang and Xu, Chao and Yi, Li and Huang, Siyuan and Wang, He},
  booktitle={Proceedings of the IEEE/CVF Conference on Computer Vision and Pattern Recognition},
  pages={7081--7091},
  year={2023}
}

@inproceedings{deng20213d,
  title={3d affordancenet: A benchmark for visual object affordance understanding},
  author={Deng, Shengheng and Xu, Xun and Wu, Chaozheng and Chen, Ke and Jia, Kui},
  booktitle={proceedings of the IEEE/CVF conference on computer vision and pattern recognition},
  pages={1778--1787},
  year={2021}
}

@article{lin2025objaverseplusplus,
  title={Objaverse++: Curated 3D Object Dataset with Quality Annotations},
  author={Lin, Chendi and Liu, Heshan and Lin, Qunshu and Bright, Zachary and Tang, Shitao and He, Yihui and Liu, Minghao and Zhu, Ling and Le, Cindy},
  journal={arXiv preprint arXiv:2504.07334},
  year={2025}
}

@article{lu2025objaverseoa,
  title={Orientation Matters: Making 3D Generative Models Orientation-Aligned},
  author={Lu, Yichong and Tian, Yuzhuo and Jiang, Zijin and Zhao, Yikun and Yang, Yuanbo and Ouyang, Hao and Hu, Haoji and Yu, Huimin and Shen, Yujun and Liao, Yiyi},
  journal={arXiv preprint arXiv:2506.08640},
  year={2025}
}

@inproceedings{jin2025canoobjdataset,
  title={One-shot 3D Object Canonicalization based on Geometric and Semantic Consistency},
  author={Jin, Li and Wang, Yujie and Chen, Wenzheng and Dai, Qiyu and Gao, Qingzhe and Qin, Xueying and Chen, Baoquan},
  booktitle={Proceedings of the Computer Vision and Pattern Recognition Conference},
  pages={16850--16859},
  year={2025}
}

@article{qian2024objaversemix,
  title={Pushing auto-regressive models for 3d shape generation at capacity and scalability},
  author={Qian, Xuelin and Wang, Yu and Luo, Simian and Zhang, Yinda and Tai, Ying and Zhang, Zhenyu and Wang, Chengjie and Xue, Xiangyang and Zhao, Bo and Huang, Tiejun and others},
  journal={arXiv preprint arXiv:2402.12225},
  year={2024}
}

@article{diazzi2021convexmeshing,
  title={Convex polyhedral meshing for robust solid modeling},
  author={Diazzi, Lorenzo and Attene, Marco},
  journal={ACM Transactions on Graphics (TOG)},
  volume={40},
  number={6},
  pages={1--16},
  year={2021},
  publisher={ACM New York, NY, USA}
}

@inproceedings{chen2025dora,
  title={Dora: Sampling and benchmarking for 3d shape variational auto-encoders},
  author={Chen, Rui and Zhang, Jianfeng and Liang, Yixun and Luo, Guan and Li, Weiyu and Liu, Jiarui and Li, Xiu and Long, Xiaoxiao and Feng, Jiashi and Tan, Ping},
  booktitle={Proceedings of the Computer Vision and Pattern Recognition Conference},
  pages={16251--16261},
  year={2025}
}

@misc{hunyuan3d_online,
      title={Hunyuan3D 3.0},
      author={Tencent},
      howpublished={\url{https://3d.hunyuan.tencent.com/}},
      note={Accessed: 2026-01-14},
      year={2026}
}
\bibliographystyle{colm2024_conference}

\end{document}